\documentclass{article}

\usepackage{PRIMEarxiv}

\usepackage[utf8]{inputenc} 
\usepackage[T1]{fontenc}    
\usepackage{hyperref}       
\usepackage{url}            
\usepackage{booktabs}       
\usepackage{amsfonts}       
\usepackage{nicefrac}       
\usepackage{microtype}      
\usepackage{lipsum}
\usepackage{fancyhdr}       
\usepackage{graphicx}       
\graphicspath{{media/}}     

\usepackage{lipsum}
\usepackage{enumitem}
\usepackage{bbold}
\newcommand{\indep}{\perp \!\!\! \perp}
\usepackage{lipsum}
\usepackage{amsmath} 

\usepackage{accents}
\newlength{\dhatheight}

\DeclareMathOperator*{\argmax}{arg\,max}

\usepackage{algorithm}
\usepackage{algpseudocode}

\newcommand{\Break}{\State \textbf{break} }
\usepackage{makecell}
\usepackage{longtable}

\title{Masked Generative Modeling with Enhanced Sampling Scheme}

\author{
  Daesoo Lee \\
  Norwegian University of Science and Technology \\
  \And
  Erlend Aune \\
  Norwegian University of Science and Technology \\ 
  BI Norwegian Business School \\
  Abelee \\ 
  \And
  Sara Malacarne \\
  Telenor Research \\
}

\begin{document}
\maketitle


\begin{abstract}
This paper presents a novel sampling scheme for masked non-autoregressive generative modeling. We identify the limitations of TimeVQVAE, MaskGIT, and Token-Critic in their sampling processes, and propose Enhanced Sampling Scheme (ESS) to overcome these limitations. 
ESS explicitly ensures both sample diversity and fidelity, and consists of three stages: Naive Iterative Decoding, Critical Reverse Sampling, and Critical Resampling. 
ESS starts by sampling a token set using the naive iterative decoding as proposed in MaskGIT, ensuring sample diversity. Then, the token set undergoes the critical reverse sampling, masking tokens leading to unrealistic samples. After that, critical resampling reconstructs masked tokens until the final sampling step is reached to ensure high fidelity. 
Critical resampling uses confidence scores obtained from a self-Token-Critic to better measure the realism of sampled tokens, while critical reverse sampling uses the structure of the quantized latent vector space to discover unrealistic sample paths. 
We demonstrate significant performance gains of ESS in both unconditional sampling and class-conditional sampling using all the 128 datasets in the UCR Time Series archive.
Code for reproducing results is available at \url{https://github.com/available-on-accept}
\end{abstract}


\section{Introduction}

In recent years, generative modeling has made significant advancements. The mainstream frameworks for generative image modeling have evolved through various phases, initially utilizing Variational AutoEncoder (VAE) \cite{kingma2013auto}, then progressing to Generative Adversarial Network (GAN) \cite{goodfellow2014generative}, and eventually Vector Quantized-Variational AutoEncoder (VQ-VAE) \cite{van2017neural} and diffusion models \cite{ho2020denoising}. 
Masked generative modeling on VQ-tokens and diffusion models have demonstrated state-of-the-art (SOTA)  performance in the last couple of years on image modeling, audio modeling, and time series modeling \cite{yu2022scaling, saharia2022photorealistic, ramesh2022hierarchical, borsos2023audiolm, liu2023audioldm, lee2023vector}.

\begin{figure}[ht]
\centering
\includegraphics[width=0.9\linewidth]{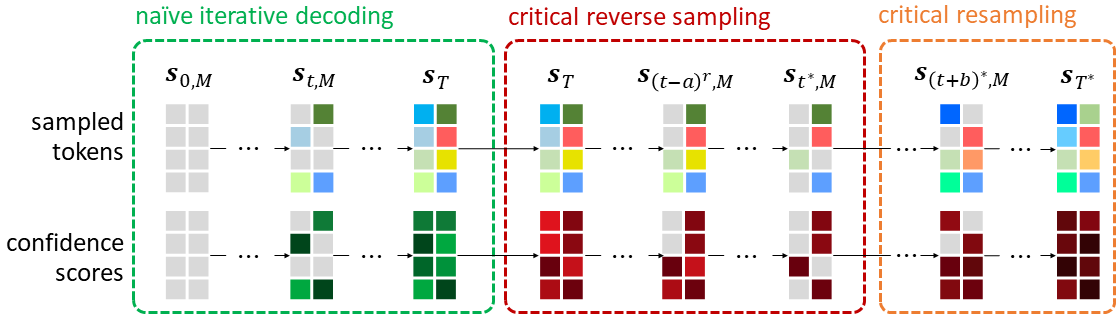}
\caption{Overview of the proposed enhanced sampling scheme. 
It consists of three stages: naive iterative decoding, critical reverse sampling, and critical resampling. 
The confidence scores $\mathcal{C}$ represent the realism of sampled tokens, and the darker color represents a higher confidence score.
Unlike the naive iterative decoding where $\mathcal{C}$ is obtained directly from the prior model, critical reverse sampling and critical resampling utilize $\mathcal{C}$ obtained with \textit{self-Token-Critic}, our proposal to better measure the realism.
In the naive iterative decoding, a token set is sampled, starting from a set of mask tokens as in MaskGIT. The sampled token set, however, contains unrealistic tokens due to the limitations of the naive iterative decoding. The critical reverse sampling masks such tokens, leaving realistic tokens only. Finally, those masked tokens are resampled via the critical resampling, producing a set of realistic tokens.}
\label{fig:overview_enhanced_sampling_scheme}
\end{figure}


Both masked generative modeling and diffusion models utilize sampling schemes to iteratively unmask/predict tokens or denoise noisy inputs. In recent years, diffusion models' sampling schemes have garnered considerable attention \cite{salimans2022progressive,song2020denoising}. While the sampling scheme for masked generative modeling has not received the same level of attention, notable progress has been made in \cite{chang2022maskgit, lezama2022improved, lee2023vector}.
This paper identifies certain limitations of the masked generative modeling sampling schemes in \cite{chang2022maskgit, lezama2022improved, lee2023vector}, and introduces a novel and improved sampling scheme which we call Enhanced Sampling Scheme (ESS) to address these limitations. ESS comprises three stages: 1) Iterative non-autoregressive decoding, as proposed in \cite{chang2022maskgit}, 2) Critical Reverse Sampling, and 3) Critical Resampling using self-Token-Critic. Figure~\ref{fig:overview_enhanced_sampling_scheme} provides an overview of the method, and Sect.~\ref{sec:method} presents a detailed explanation of ESS. Notably, our proposed approach can be seamlessly integrated into any masked generative model relying on VQVAE without requiring any modifications to its training process. Instead, ESS solely replaces the existing sampling process. 

We demonstrate the efficacy of ESS on all the 128 datasets in the UCR Time Series archive \cite{UCRArchive2018}. This collection of datasets has high diversity with respect to their distributions, making it a solid testbed for evaluating the performance of ESS. We leave further evaluations of ESS in other domains such as image modeling or audio modeling for a future study.

Our experimental results clearly indicate that ESS generates time series with substantially improved fidelity and increased sample diversity compared to the baselines.
Furthermore, our ablation study demonstrates that our proposed approach achieves more effective sampling than Token-Critic \cite{lezama2022improved}, without the need for training an auxiliary critic model.

To summarize, our contributions consist of 
\begin{enumerate}[topsep=0pt,itemsep=-1ex,partopsep=1ex,parsep=1ex]
    \item[$\bullet$] Identification of limitations of MaskGIT, TimeVQVAE and Token-Critic,
    \item[$\bullet$] A new sampling scheme, ESS, that improves the sampling process for masked generative modeling,
    \item[$\bullet$] SOTA time series generation performance.
\end{enumerate}

\section{Background and Related work}

VQ-VAE introduced a two-staged training approach: the first stage focuses on learning the projection of input data to a discrete latent space and reconstructing it back to the data space. More specifically, in this stage, the data is first projected to a latent space which is then discretized using vector quantization. Each discrete vector is called token. The second stage involves training a prior model to capture the prior distribution of the discrete tokens. An autoregressive transformer is deployed for the prior learning stage, it is trained to predict the next token given the previous tokens. 

The follow-up papers by \cite{chang2022maskgit, lezama2022improved} have further improved the image generation performance. \cite{chang2022maskgit} identified limitations in the second stage of VQ-VAE, particularly regarding its autoregressive generative modeling. To address this, the authors proposed a bidirectional generative modeling approach called MaskGIT, which enhances both fidelity and sampling speed.
Later, \cite{lezama2022improved} discovered certain flaws in MaskGIT and proposed a third stage to improve its sampling process. In this third stage, a discriminator called Token-Critic is trained to distinguish between real samples and generated samples. The discriminative scores obtained from Token-Critic are then utilized in MaskGIT's sampling process, further enhancing the quality of the generated samples.

Motivated by the success in generative image modeling, recent work in \cite{lee2023vector} introduced TimeVQVAE as the first approach to leverage VQ for time series generation (TSG), demonstrating SOTA performance in TSG \cite{ang2023tsgbench}. Specifically, TimeVQVAE employs a two-stage approach, utilizing a modified time-frequency-based VQ-VAE for the first stage and MaskGIT for the second stage. Prior to the development of TimeVQVAE, the field of TSG had primarily been dominated by GANs \cite{esteban2017real, yoon2019time, ni2020conditional, smith2020conditional, li2022tts}.

\begin{figure*}[!ht]
\centering
\includegraphics[width=0.999\linewidth]{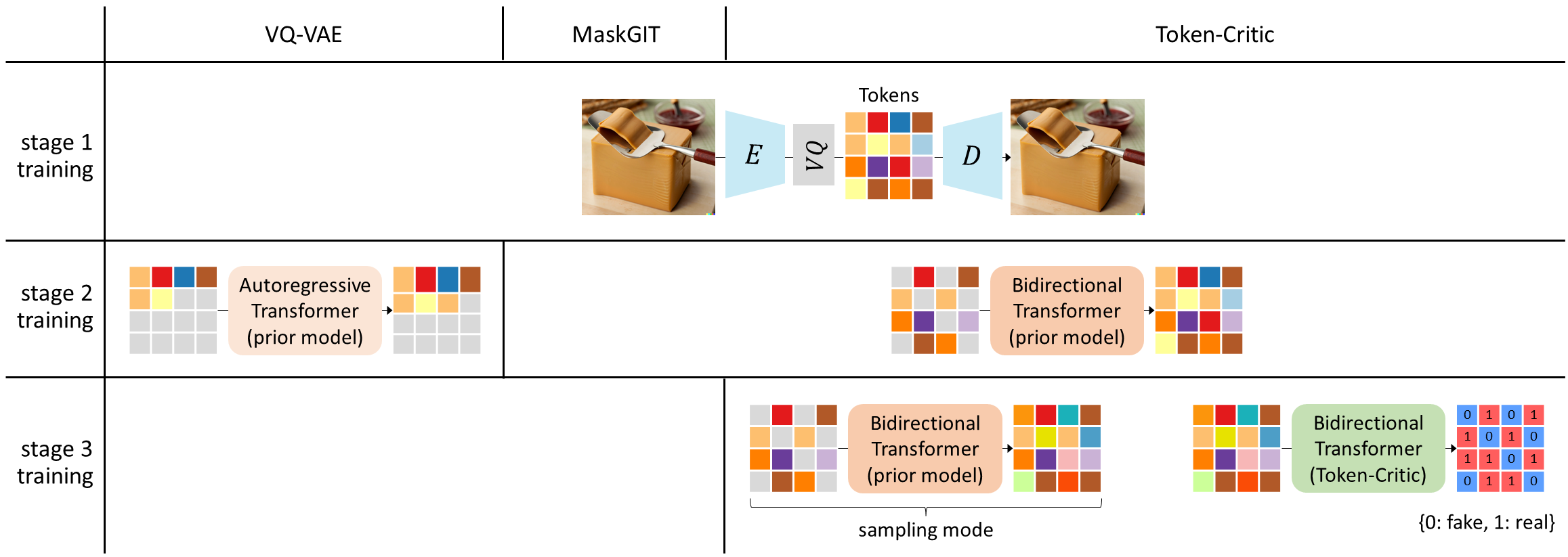}
\caption{Overviews of the training stages in VQ-VAE, MaskGIT, and Token-Critic.
The first stage (stage~1) is shared between VQ-VAE, MaskGIT, and Token-Critic, and the second stage (stage~2) of VQ-VAE is different from MaskGIT and Token-Critic, while MaskGIT and Token-Critic have the same stage~2. Lastly, the third stage (stage~3) is only present in Token-Critic.}
\label{fig:overview_vqvae_maskgit_tokencritic}
\end{figure*}

\begin{figure}[!ht]
\centering
\includegraphics[width=0.6\linewidth]{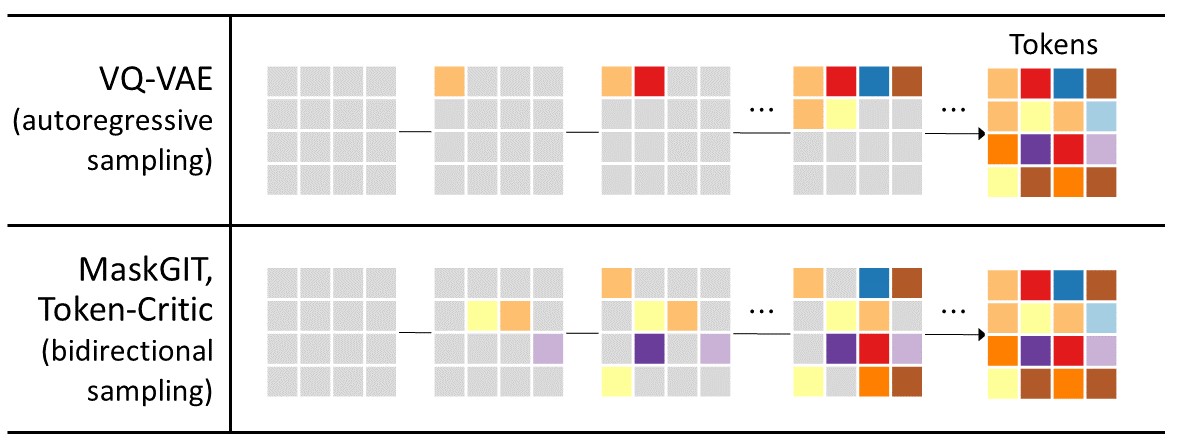}
\caption{Overviews of the sampling process in VQ-VAE, MaskGIT, and Token-Critic.
Tokens are autoregressively sampled in VQ-VAE, while bidirectionally sampled in MaskGIT and Token-Critic.
Notably, the bidirectional sampling samples multiple tokens at a time, allowing for faster sampling.}
\label{fig:overview_sampling_process}
\end{figure}

Figure~\ref{fig:overview_vqvae_maskgit_tokencritic} illustrates the training stages in VQ-VAE, MaskGIT, and Token-Critic. 
We denote, $E$, $VQ$, and $D$ as encoder, vector quantizer, and decoder, respectively. A token refers to the codebook index of a discrete latent vector \cite{chang2022maskgit}, where the discrete latent vector is the output after  $E$ and $VQ$. Each color corresponds to a different token while the grey token refers to a masked token \cite{chang2022maskgit}. In stage~3, the sampling mode indicates that the prior model samples new tokens to replace the grey tokens in the output, and Token-Critic receives the sampled tokens as input and outputs the discriminative scores. 
Furthermore, figure~\ref{fig:overview_sampling_process} illustrates the sampling processes in VQ-VAE, MaskGIT, and Token-Critic.

\subsection{MaskGIT}
Stage~1 remains the same as VQ-VAE (we refer to \cite{van2017neural} for a detailed description), while Stage~2 somewhat deviates from that of VQ-VAE. MaskGIT tackles the limitation of VQ-VAE posed by the autoregressive nature of its prior model by introducing a bidirectional transformer for a prior model. The model is trained by making predictions on a set of randomly-masked tokens. For details, see \cite{chang2022maskgit, lee2023vector}.

\begin{figure}[!ht]
\centering
\includegraphics[width=0.45\linewidth]{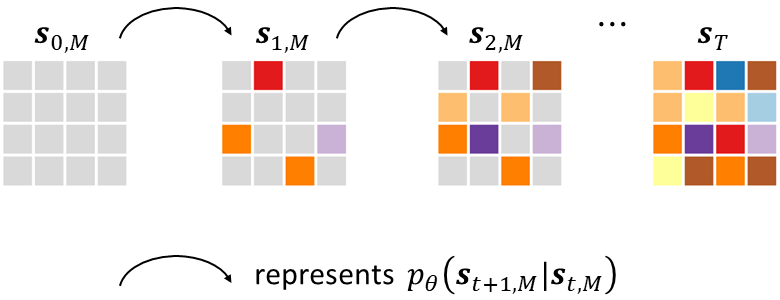}
\caption{The iterative sampling process of MaskGIT, named iterative decoding.}
\label{fig:iteartive_decoding_maskgit}
\end{figure}

When sampling, MaskGIT uses an iterative scheme called iterative decoding. This scheme was proposed in \cite{chang2022maskgit}, where masked tokens are iteratively unmasked using the learned prior model. The set of sampled tokens is decoded back to the data space using $D$. 
Figure~\ref{fig:iteartive_decoding_maskgit} illustrates the sampling process of MaskGIT. $\textit{\textbf{s}}_{t,M}$ denotes a set of masked tokens at the decoding step $t$. The decoding steps range from 0 to $T$, in which $\textit{\textbf{s}}_T$ is the final outcome in the sampling process. $\theta$ denotes the learnable parameters of the prior model. The single decoding process $p_\theta(\textit{\textbf{s}}_{t+1, M} | \textit{\textbf{s}}_{t,M})$ consists of two sub-processes: $p_\theta(\textit{\textbf{s}}_t | \textit{\textbf{s}}_{t, M})$ and $p(\textit{\textbf{s}}_{t+1, M} | \textit{\textbf{s}}_t)$, forming $p_\theta(\textit{\textbf{s}}_{t+1, M} | \textit{\textbf{s}}_{t,M}) = \sum_{\textit{\textbf{s}}_t} p(\textit{\textbf{s}}_{t+1, M} | \textit{\textbf{s}}_t)  p_\theta(\textit{\textbf{s}}_t | \textit{\textbf{s}}_{t, M})$.
In the equation, $p_\theta(\textit{\textbf{s}}_t | \textit{\textbf{s}}_{t, M})$ refers to sampling an entire set of tokens using the prior model, and $p(\textit{\textbf{s}}_{t+1, M} | \textit{\textbf{s}}_t)$ refers to masking the sampled tokens with low probabilities 
for the next decoding step according to a masking scheduler. We represent the scores that assess the realism of sampled tokens as confidence scores denoted by $\mathcal{C}$.

To be more concrete, the sampling process of iterative decoding can be formulated as 
\begin{equation}
\label{eq:naive_iterative_decoding_sampling_step}
\begin{split}
p_\theta({\textit{\textbf{s}}}_{t+1, M} | {\textit{\textbf{s}}}_{t, M}) & = \sum_{{\textit{\textbf{s}}}_t} p({\textit{\textbf{s}}}_{t+1, M} | {\textit{\textbf{s}}}_t) p_\theta({\textit{\textbf{s}}}_t | {\textit{\textbf{s}}}_{t, M}) \\
& = \sum_{{\textit{\textbf{s}}}_t} p({\textit{\textbf{s}}}_{t+1, M} | {\textit{\textbf{s}}}_t) \prod_{i} p_\theta ((\textit{\textbf{s}}_t)_i | {\textit{\textbf{s}}}_{t, M})
,
\end{split}
\end{equation}
where $(\textit{\textbf{s}}_t)_i$ is the $i$-th element of $\textit{\textbf{s}}_t$. 
We compute ${\textit{\textbf{s}}}_{t+1, M}$ using the expression $\textit{\textbf{m}}_{t} \odot {\textit{\textbf{s}}}_{t}$, where $\textit{\textbf{m}}_t$ denotes a masking matrix consisting of 0 and 1, indicating the positions to be masked and unmasked, respectively. The complete-masking matrix, denoted by $\textit{\textbf{m}}_0$, consists of only zeros, while the completely-unmasking matrix, denoted by $\textit{\textbf{m}}_T$, consists of only ones. 
In relation to the confidence scores, $\textit{\textbf{m}}_t$ is computed as
\begin{equation}
\label{eq:m_t_naive_iterative_decoding}
(\textit{\textbf{m}}_t)_i = 
\begin{cases}
1, & \parbox{0.6\linewidth}{if the confidence score $\mathcal{C}_i$ is among the top \textit{K}}  \\
0, & \text{otherwise}
\end{cases}
\end{equation}
where $(\textit{\textbf{m}}_t)_i \in \textit{\textbf{m}}_t$ and $\mathcal{C}_i = p_\theta ((\textit{\textbf{s}}_t)_i | {\textit{\textbf{s}}}_{t, {M}})$. 
Notably, the tokens sampled at $t$ using $\textit{\textbf{m}}_t$ cannot be corrected at a later step.

\subsection{Token-Critic}
Stage~1 and Stage~2 remain the same as MaskGIT. In Stage~3, the key difference is the presence of a discriminator named Token-Critic trained to differentiate between real and generated tokens. 
The input of Token-Critic is a random-combination of real and generated tokens, sampled using $p_\theta(\textit{\textbf{s}} | \textit{\textbf{s}}_M)$ where $\textit{\textbf{s}}_M$ refers to a set of randomly-masked tokens.
Using the input, the Token-Critic is trained to output a discriminative score of 0 for generated tokens (\textit{i.e.,} fake tokens) and 1 for real tokens. 

The discriminative scores are utilized in the masking process of  MaskGIT's sampling process, specifically in $p(\textit{\textbf{s}}_{t+1, M} | \textit{\textbf{s}}_t)$. Unlike \cite{chang2022maskgit}, \cite{lezama2022improved} masks the sampled tokens with low values of the discriminative scores, equivalently masking the likely-fake tokens. Hence, the confidence scores in Token-Critic are the discriminative scores. Their experimental results demonstrated the improvement in fidelity and sample diversity compared to the MaskGIT's iterative decoding.

More precisely, Token-Critic reformulates $p({\textit{\textbf{s}}}_{t+1, M} | {\textit{\textbf{s}}}_t)$ as $p_\phi({\textit{\textbf{s}}}_{t+1, M} | {\textit{\textbf{s}}}_t)$ where $\phi$ denotes parameters of Token-Critic. It computes $\textit{\textbf{m}}_t$ in the same way as \eqref{eq:m_t_naive_iterative_decoding}. 
The main distinction lies in the calculation of the confidence score, which is obtained as $\mathcal{C}_i = g_\phi({\textit{\textbf{s}}}_t)$ for all $i$, where $g_\phi$ denotes Token-Critic. Following the same notation, calculation of $\mathcal{C}_i$ in MaskGIT can be rewritten as $\mathcal{C}_i = f_\theta({\textit{\textbf{s}}}_{t, M})$ where $f_\theta$ denotes the prior model. 
Then it becomes apparent that the capability of Token-Critic comes from being able to model $p ((\textit{\textbf{s}}_t)_i | {\textit{\textbf{s}}}_{t})$ rather than $p ((\textit{\textbf{s}}_t)_i | {\textit{\textbf{s}}}_{t, M})$.

\subsection{Limitations of VQ-VAE, MaskGIT, and Token-Critic}
\label{sect:limitations}



The generative capabilities of VQ-VAE are constrained by the nature of an autoregressive model in stage~2. One of the inherent limitations of autoregressive modeling is the sampling speed of the models: being able to sample only one token per model evaluation rather than multiple as in MaskGIT severely slows down the generative process. Autoregressive sampling also has the limitation that it may not sample the "optimal" next token given the set of unmasked tokens: it is not given that the \textit{next} token is the \textit{best} token for generative modeling.
This poses challenges for modeling long-term correlation and significantly slows down the sampling process.

MaskGIT resolved the VQ-VAE's limitation by introducing a bidirectional transformer with the sampling process described above. However, it was found by \cite{lezama2022improved} that the iterative decoding sampling has several drawbacks. 
First, its token sampling process for each token is independent of other tokens, that is $p_\theta((\textit{\textbf{s}}_t)_i | \textit{\textbf{s}}_{t,M}) \indep p_\theta((\textit{\textbf{s}}_t)_j | \textit{\textbf{s}}_{t,M})$ where $i \neq j$ (independent sampling), resulting in the reduction of fidelity of generated samples. The independent sampling compromises the capture of rich correlations between tokens. Second, it does not allow for correcting the tokens sampled in the previous decoding steps in the subsequent steps (uncorrectable sampling). Lastly, the token sampling relies on the prior model's predicted probability, which may be susceptible to modeling errors (modeling error).

Token-Critic addresses the three limitations of MaskGIT's iterative decoding. The issues of independent sampling and uncorrectable sampling are overcome by measuring the confidence score of each sampled token considering the other sampled tokens via $\mathcal{C} = g_\phi(\textit{\textbf{s}}_t)$. 
The challenge of modeling error is mitigated through the introduction of Token-Critic, specifically designed to assess the realism of sampled tokens.

Despite its strengths, Token-Critic is not without flaws. We found that the discriminative training of Token-Critic struggles when real tokens and generated samples are difficult to be distinguished. This occurs when the prior model has been effectively trained, thereby being able to sample realistic tokens. In such a scenario, we observed that its training loss, a binary cross-entropy loss, does not decrease much. As a result, its discriminative scores do not properly capture the realism of generated samples, limiting the capability of the sampling process. In addition, Token-Critic requires an additional training stage, therefore it adds more complexity and computational costs.

\subsection{TimeVQVAE}
\begin{figure}[!ht]
\centering
\includegraphics[width=0.75\linewidth]{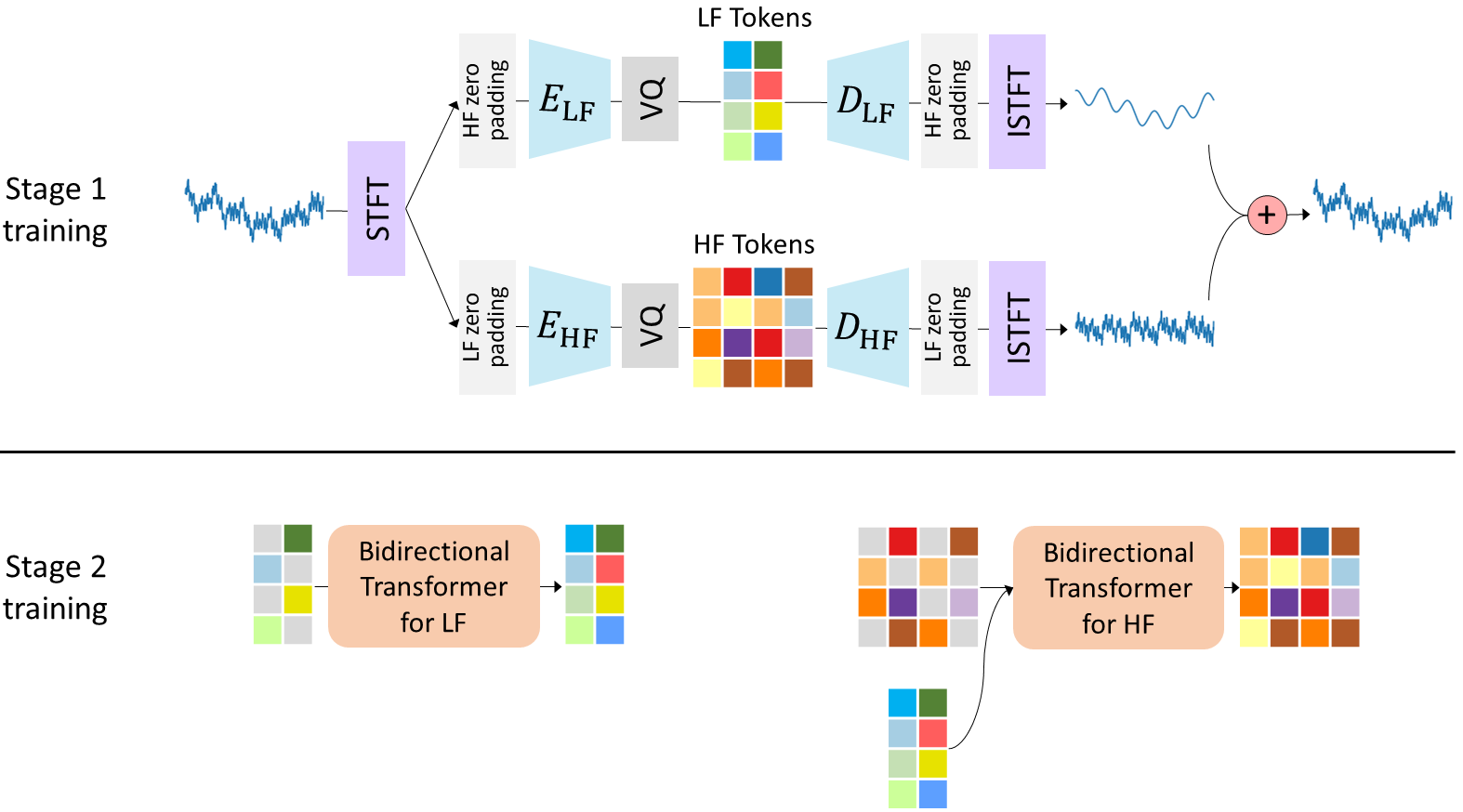}
\caption{Overview of TimeVQVAE.}
\label{fig:overview_timevqvae}
\end{figure}

The overview of TimeVQVAE is illustrated in figure~\ref{fig:overview_timevqvae}, where LF and HF denote low-frequency and high-frequency, STFT and ISTFT denote Short Term Fourier Transform and its inverse process, respectively, and HF and LF zero padding denote zero padding on the HF and LF regions of $\mathrm{STFT}(x)$, respectively. See \cite{lee2023vector} for details. While the overall framework of TimeVQVAE is somewhat similar to the above-mentioned image generative modeling approaches, the key difference resides in the latent space separation with respect to two different frequency bands. The separation breaks the TSG problem into two smaller tasks, easing the generative modeling.

Similar to VQ-VAE, the stage~1 training is conducted by minimizing a reconstruction loss and an additional loss to connect a gradient flow between $E$ and $VQ$.
In stage~2, training a prior model for LF is the same as MaskGIT, but training a prior model for HF is different, as it solves $p(\textit{\textbf{s}}^{\mathrm{HF}} | \textit{\textbf{s}}^\mathrm{HF}_M, \textit{\textbf{s}}^\mathrm{LF})$.

The sampling is performed similarly to MaskGIT but in two steps: first, LF sampling and subsequently HF sampling. The LF sampling process is identical to MaskGIT's, while the HF sampling process is conditioned on the sampled LF tokens.

TimeVQVAE carries the same limitations of MaskGIT such as the independent sampling, uncorrectable sampling, and modeling error. 
We experimentally found that the application of Token-Critic to TimeVQVAE somewhat improves the generative performance but the improvement is restrained by the limitations of Token-Critic.
In this paper, we propose an enhanced sampling scheme to be free of such limitations and in turn achieve a superior TSG performance.

\section{Method: Enhanced Sampling Scheme} \label{sec:method}

Our proposed enhanced sampling scheme consists of three stages: 1) naive iterative decoding sampling process, as in MaskGIT, 2) removal of less likely tokens, named  \textit{Critical Reverse Sampling}, 3) resampling for those tokens that have been removed, named  \textit{Critical Resampling}. 
Notably, the second and third stages use a \textit{confidence} score obtained with \textit{self-Token-Critic} -- that is, our proposal to better measure the realism of sampled tokens without requiring either an additional training stage or model. 

Figure~\ref{fig:overview_enhanced_sampling_scheme} illustrates the overview of the enhanced sampling scheme. The green confidence scores are computed by $p_\theta(\textit{\textbf{s}}_t | \textit{\textbf{s}}_{t,M})$, same as in the naive iterative decoding, the red confidence scores are computed using self-Token-Critic. The darker color represents a higher confidence score. $a$ and $b$ are strictly positive integers, $t^r$ denotes a step during the critical reverse sampling, $t^*$ denotes the step where the critical reverse sampling has stopped, and $T^*$ denotes the last step of the critical resampling. 
It should be noted that the confidence scores measured by $p_\theta(\textit{\textbf{s}}_t | \textit{\textbf{s}}_{t,M})$ and self-Token-Critic should be different to some degree, as our self-Token-Critic measures the realism of sampled tokens better, overcoming the modeling error problem, discussed in Sect.~\ref{sect:limitations}. Furthermore, self-Token-Critic enables correctable sampling, which is not available through the naive iterative decoding, as also discussed in Sect.~\ref{sect:limitations}.
The sampling process starts by sampling a token set using the same method as MaskGIT. Subsequently, the token set undergoes the critical reverse sampling until the less likely tokens are masked. Then, the critical resampling takes the reversed token set and resamples tokens until the final step $T^*$ is reached. The resulting token set is then decoded to a time series.


\subsection{Self-Token-Critic}
Token-Critic, despite its strengths, has some limitations. It struggles to distinguish between real and generated tokens when the prior model produces realistic tokens. As a result, its training loss does not decrease significantly, negatively affecting the accuracy of discriminative scores. Additionally, Token-Critic requires an extra training stage, adding complexity and computational costs. 

Self-Token-Critic is proposed to approximate Token-Critic so that we preserve its benefits, such as resolving the independent sampling, uncorrectable sampling, and the modeling error problems, while discarding its drawbacks. 

Motivated by the findings that the strength of Token-Critic stems from being able to model $p((\textit{\textbf{s}}_t)_i | \textit{\textbf{s}}_t)$, we approximate it by reformulating the equation as
\begin{subequations}
\label{eq:approx_token_critic}
\begin{align}
&p ((\textit{\textbf{s}}_t)_i | {\textit{\textbf{s}}}_{t}) \\ & = p ((\textit{\textbf{s}}_t)_i | (\textit{\textbf{s}}_t)_{0}, (\textit{\textbf{s}}_t)_{1}, \cdots, (\textit{\textbf{s}}_t)_{i-1}, (\textit{\textbf{s}}_t)_i, (\textit{\textbf{s}}_t)_{i+1}, \cdots) \\
& \approx p ((\textit{\textbf{s}}_t)_i | (\textit{\textbf{s}}_t)_{0}, (\textit{\textbf{s}}_t)_{1}, \cdots, (\textit{\textbf{s}}_t)_{i-1}, (\textit{\textbf{s}}_t)_{i+1}, \cdots) \\
& = p ((\textit{\textbf{s}}_t)_i | {\textit{\textbf{s}}}_{t, M_i}), 
\end{align}
\end{subequations}
where $M_i$ denotes masking the $i$-th element and the approximation is reasonably accurate if ${\textit{\textbf{s}}}_{t}$ is sufficiently long. 
The approximated expression $p ((\textit{\textbf{s}}_t)_i | {\textit{\textbf{s}}}_{t, M_i})$ is a special case of the naive iterative decoding step expressed as $p ((\textit{\textbf{s}}_t)_i | {\textit{\textbf{s}}}_{t, M})$, where only the $i$-th token is masked. As a result, we could compute the confidence score $\mathcal{C}_i$ using the existing prior model as $p_\theta((\textit{\textbf{s}}_t)_i | {\textit{\textbf{s}}}_{t, M_i})$.
This allows us to keep the benefits of Token-Critic without requiring an additional model or training stage.

However, a direct use of the confidence score obtained with the prior model is subject to the modeling error problem, as the prior model's output probability may be unreliable. This can lead to sub-optimal predictions of the confidence scores and eventually sub-optimal fidelity. To address this, we compute the confidence scores in the latent space learned in stage~1, where higher-level semantics of time series patterns are effectively captured as discussed in Sect.~\ref{sect:critical_reserve_sampling}. 
Specifically, the confidence score $\mathcal{C}_i$ of $\textit{\textbf{s}}_t$ can be expressed as
\begin{subequations}
\label{eq:confdience_score}
\begin{align}
\mathcal{C}_i &= \frac{e^{d_i}}{\sum_j e^{d_j}} \;\; \forall i \\
d_j &= - \sum{((\textit{\textbf{z}}^q_t)_j - (\tilde{\textit{\textbf{z}}}^q_t)_j)^2} \;\; \forall j \\
(\textit{\textbf{z}}^q_t)_j &= \text{discrete latent vector of }(\textit{\textbf{s}}_t)_j \\
(\tilde{\textit{\textbf{z}}}^q_t)_j &= \text{discrete latent vector of }(\tilde{\textit{\textbf{s}}}_t)_j \\
(\tilde{\textit{\textbf{s}}}_t)_j &= \text{argmax } p_\theta(\textit{\textbf{s}}_j | \textit{\textbf{s}}_{t, M_j}) p(\textit{\textbf{s}}_{t, M_j} | \textit{\textbf{s}}_t),
\end{align}
\end{subequations}
where $d_j$ measures how far the sampled token is located apart from the most likely token in the latent space learned in stage 1, equivalently measuring the realism of the sampled token and $p(\textit{\textbf{s}}_{t, M_j} | \textit{\textbf{s}}_t)$ is a deterministic process of masking a single token.

\subsection{Critical Reverse Sampling}
\label{sect:critical_reserve_sampling}
The set of sampled tokens from the naive iterative decoding inevitably contains less likely tokens due to the independent sampling, as discussed in Sect.~\ref{sect:limitations}. Our motivation behind the critical reverse sampling is to effectively remove the less likely tokens from the sampled token set, $\textit{\textbf{s}}_{T,M}$ by retracting the decoding step to $t^*$ where the less likely tokens are all removed. To do that, we need to determine its parameter $t^*$. 

Determining the optimal value of $t^*$, however, can be challenging. The ideal value of $t^*$ varies depending on the number of less likely tokens in a token set. Choosing an arbitrary value for $t^*$ as a hyperparameter leads to an ad-hoc approach, making it necessary to find a better way.

We propose to choose $t^*$ by utilizing a transitional direction of ${\textit{\textbf{s}}}_{t,M}$, expressed as ${\partial {\textit{\textbf{s}}}_{t,M}} = {\textit{\textbf{s}}}_{t,M} - {\textit{\textbf{s}}}_{t-1,M}$, where ${\textit{\textbf{s}}}_{t,M}$ and  ${\textit{\textbf{s}}}_{t-1,M}$ can be easily obtained by masking ${\textit{\textbf{s}}}_{T}$ according to the confidence scores obtained with self-Token-Critic and a masking scheduler. 
${\partial {\textit{\textbf{s}}}_{t,M}}$ equivalently represents the additional tokens existing in ${\textit{\textbf{s}}}_{t,M}$ compared to ${\textit{\textbf{s}}}_{t-1,M}$.
Given that ${\textit{\textbf{s}}}_{t,M}$ contains less likely tokens, ${\partial {\textit{\textbf{s}}}_{t,M}}$ indicates a direction towards a sub-realistic token set. 
The direction towards a realistic token set, on the other hand, can be expressed as ${\partial \tilde{\textit{\textbf{s}}}_{t, M}} = \tilde{\textit{\textbf{s}}}_{t, M} - {\textit{\textbf{s}}}_{t-1, M}$, where $\tilde{\textit{\textbf{s}}}_{t,M} = \text{argmax } p ({\textit{\textbf{s}}}_{t, M} | {\textit{\textbf{s}}}_{t-1}) p_\theta({\textit{\textbf{s}}}_{t-1} | {\textit{\textbf{s}}}_{t-1, M})$. 
Then, the similarity comparison between ${\partial {\textit{\textbf{s}}}_{t,M}}$ and ${\partial \tilde{\textit{\textbf{s}}}_{t,M}}$ can tell us if ${\textit{\textbf{s}}}_{t,M}$ is on a path towards a realistic token set or not. 

To summarize, the critical reverse sampling iteratively retracts the decoding step from $T$ while masking the \textit{n} least realistic tokens until the difference between ${\partial {\textit{\textbf{s}}}_{t,M}}$ and ${\partial \tilde{\textit{\textbf{s}}}_{t,M}}$ is small enough, finally reaching $t^*$. 

\begin{figure}[!ht]
\centering
\includegraphics[width=0.6\linewidth]{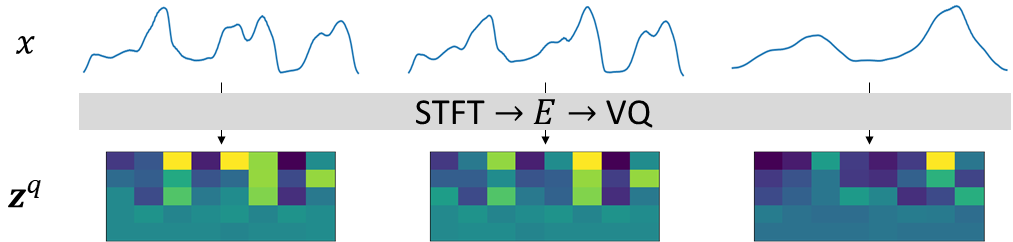}
\caption{Example of time series with both similar and dissimilar patterns compared in the discrete latent space. The first two time series have similar patterns, while the last one has a dissimilar pattern. 
Below, a matrix of codebook vectors $\textit{\textbf{z}}^q$ is visualized, where the hidden dimension is reduced to 1 using PCA for visualization. 
It demonstrates that high-level semantics of time series patterns are well captured in the discrete latent space of stage~1, allowing for effective differentiation between samples with different patterns. 
}
\label{fig:z_class0_vs_z_class1_in_stage1_latent_space}
\end{figure}

The difference between ${\partial {\textit{\textbf{s}}}_{t,M}}$ and ${\partial \tilde{\textit{\textbf{s}}}_{t,M}}$
can simply be measured in the latent space learned in stage~1. Specifically, we first obtain the corresponding discrete latent vectors of ${\partial {\textit{\textbf{s}}}_{t,M}}$ and ${\partial \tilde{\textit{\textbf{s}}}_{t,M}}$ and compute the Euclidean distance.
This is an effective approach because the discrete latent space of stage~1 captures high-level semantics of data patterns, leading to measuring the high-level semantical difference between ${\partial {\textit{\textbf{s}}}_{t,M}}$ and ${\partial \tilde{\textit{\textbf{s}}}_{t,M}}$. 
Figure~\ref{fig:z_class0_vs_z_class1_in_stage1_latent_space} presents an example of time series with both similar and dissimilar patterns along with the similarity and dissimilarity in the discrete latent space.

\subsection{Critical Resampling}
The reversed token set is resampled, iteratively moving the sampling step from $t^*$ to $T^*$ with the better confidence scores obtained by using the self-Token-Critic. 
The key difference from the naive iterative decoding is that the critical resampling allows for the dependent and correctable token sampling while easing the modeling error problem.

\subsection{Pseudocode}
The pseudocode of enhanced sampling scheme (\textit{i.e.,} naive iterative decoding $\rightarrow$ critical reverse sampling $\rightarrow$ critical resampling) is presented in Algorithm~\ref{alg:pseudocode_critical_reverse_sampling}.
There exists one threshold parameter, $\tau$.
In practice, a moving-averaged ratio of $\mathbb{E} \left[ ( \textit{\textbf{z}}^q_{t-1,M} - \tilde{\textit{\textbf{z}}}^q_{t-1,M} )^2 \right] / \, \mathbb{E} \left[ ( \textit{\textbf{z}}^q_{t,M} - \tilde{\textit{\textbf{z}}}^q_{t,M} )^2 \right] \leq 1$
can be used instead of $\mathbb{E} \left[ ( \textit{\textbf{z}}^q_{t,M} - \tilde{\textit{\textbf{z}}}^q_{t,M} )^2 \right] \leq \tau$ to discard the choice of $\tau$, making the critical reverse sampling non-parametric.
In the beginning of the critical reverse sampling, the ratio value is likely to be large, suggesting that a considerable amount of less likely tokens has been removed. As the ratio approaches 1, it indicates a flattening of the reduction in less likely tokens.

\begin{algorithm}[!ht]
\caption{Pseudocode of the enhanced sampling scheme}
\label{alg:pseudocode_critical_reverse_sampling}
\begin{algorithmic}

\For{$t \gets 0$ to $T-1$}  \Comment{naive iterative decoding}
    \State sample $\textit{\textbf{s}}_{t}$ via $\prod_{i} p_\theta ((\textit{\textbf{s}}_t)_i | {\textit{\textbf{s}}}_{t, M})$.
    \State $\mathcal{C} \gets p_\theta((\textit{\textbf{s}}_t)_i | \textit{\textbf{s}}_{t, M})$ for all $i$  
    \State assign $\infty$ to $C_i$ for the tokens that have previously been sampled to keep those tokens unmaksed.
    \State add a scheduled noise to $\mathcal{C}$ to add randomness.  \Comment{the noise magnitude decreases towards $T-1$}
    \State $\textit{\textbf{s}}_{t+1,M} \gets$ masking $\textit{\textbf{s}}_t$ given $\mathcal{C}$ and the masking scheduler  
\EndFor
\State acquire $\textit{\textbf{s}}_{T}$  \Comment{$\textit{\textbf{s}}_{T} = \textit{\textbf{s}}_{T, M}$}
\\
\State compute the confidence scores $\mathcal{C}$ of ${\textit{\textbf{s}}}_T$ using self-Token-Critic

\For{$t \gets T$ to $1$}  \Comment{critical reverse sampling}
    \State compute ${\textit{\textbf{s}}}_{t,M}$ and ${\textit{\textbf{s}}}_{t-1,M}$ by masking ${\textit{\textbf{s}}}_T$ given $\mathcal{C}$ and the masking scheduler.  
    \State compute $\tilde{\textit{\textbf{s}}}_{t,M}$ by $\argmax p ({\textit{\textbf{s}}}_{t, M} | {\textit{\textbf{s}}}_{t-1}) p_\theta({\textit{\textbf{s}}}_{t-1} | {\textit{\textbf{s}}}_{t-1, M})$
    \State compute ${\partial {\textit{\textbf{s}}}_{t,M}} = {\textit{\textbf{s}}}_{t,M} - {\textit{\textbf{s}}}_{t-1,M}$  \Comment{additional tokens existing in ${\textit{\textbf{s}}}_{t,M}$ compared to ${\textit{\textbf{s}}}_{t-1,M}$}
    \State compute ${\partial \tilde{\textit{\textbf{s}}}_{t,M}} = \tilde{\textit{\textbf{s}}}_{t,M} - {\textit{\textbf{s}}}_{t-1,M}$  \Comment{additional tokens existing in ${\tilde{\textit{\textbf{s}}}}_{t,M}$ compared to ${\textit{\textbf{s}}}_{t-1,M}$}

    \State $\textit{\textbf{z}}^q_{t,M} \gets$ discrete latent vector of $\partial {\textit{\textbf{s}}}_{t,M}$
    \State $\tilde{\textit{\textbf{z}}}^q_{t,M} \gets$ discrete latent vector of $\partial \tilde{\textit{\textbf{s}}}_{t,M}$

    \If {$\mathbb{E} \left[ ( \textit{\textbf{z}}^q_{t,M} - \tilde{\textit{\textbf{z}}}^q_{t,M} )^2 \right] \leq \tau$}  \Comment{$\tau$ denotes a threshold and the difference is measured in the discrete latent space learned in stage~1.}
        \State $t^* \gets t$
        \Break
    \EndIf
\EndFor
\State acquire $\textit{\textbf{s}}_{t^*, M}$
\\
\For{$t \gets t^*$ to $T^*-1$}  \Comment{critical resampling}
    \State sample $\textit{\textbf{s}}_{t}$ via $\prod_{i} p_\theta ((\textit{\textbf{s}}_t)_i | {\textit{\textbf{s}}}_{t, M})$.
    \State compute $\mathcal{C}$ of $\textit{\textbf{s}}_t$ using self-Token-Critic.
    \State add a scheduled noise to $\mathcal{C}$ to add randomness.
    \State $\textit{\textbf{s}}_{t+1,M} \gets$ masking $\textit{\textbf{s}}_t$ given $\mathcal{C}$ and the masking scheduler
\EndFor
\State acquire $\textit{\textbf{s}}_{T^*}$  \Comment{$\textit{\textbf{s}}_{T^*} = \textit{\textbf{s}}_{T^*, M}$}

\end{algorithmic}
\end{algorithm}

\subsection{Methodological Discussion}

\begin{figure}[!ht]
\centering
\includegraphics[width=0.8\linewidth]{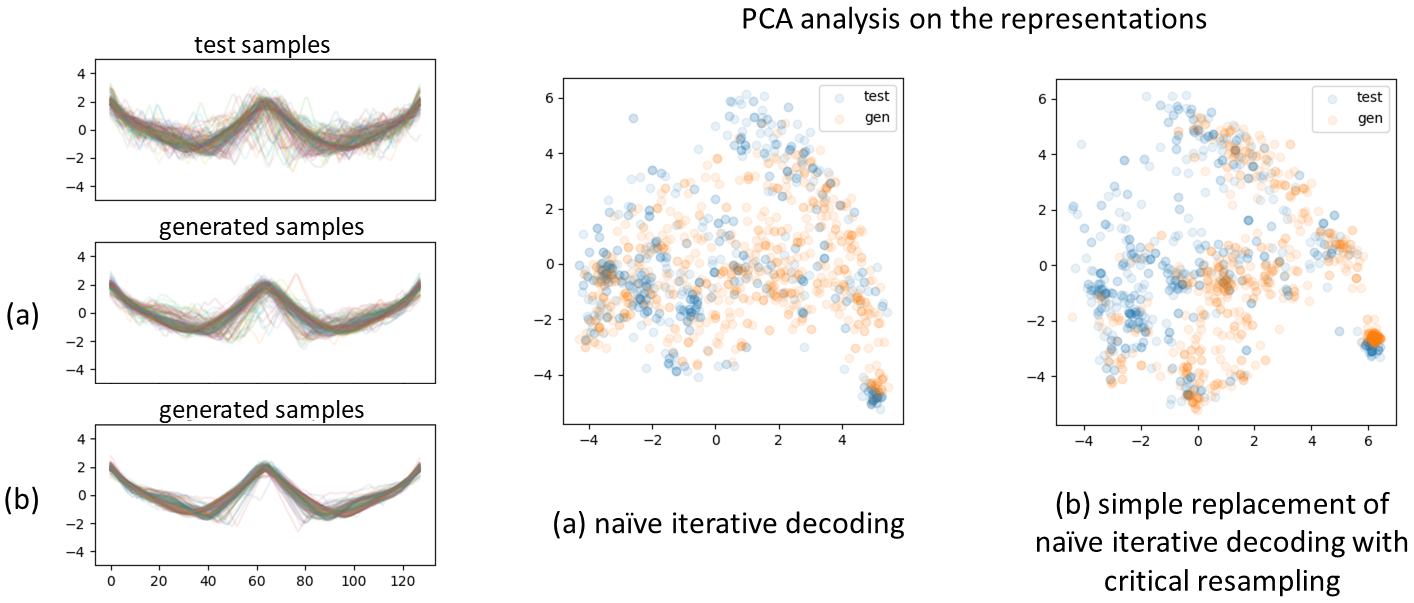}
\caption{Visualization of the representations of two sets of generated time series. The representations are acquired using the pretrained FCN model as introduced in \cite{lee2023vector} and their dimension is reduced to 2 through PCA for visualization. The blue and orange dots denote representations of test samples and generated samples, respectively.
(a) is generated using the naive iterative decoding and (b) is generated by simply replacing the naive iterative decoding with the critical resampling. 
The generated samples in (b) exhibit lower sample diversity.
}
\label{fig:token_critic_saturation_example}
\end{figure}

\paragraph{Why Not Replacing Naive Iterative Decoding with Critical Resampling?}
While it may seem straightforward to simply replace the naive iterative decoding with the critical resampling to keep the sampling process short, our experiments have shown that doing so can improve the fidelity of generated samples but compromises the sample diversity. This is because self-Token-Critic strongly guides the sampling process towards highly-likely tokens, which can limit the range of samples to be produced, as shown in figure~\ref{fig:token_critic_saturation_example}.

\paragraph{Relations between Naive Iterative Decoding, Critical Reverse Sampling, and Critical Resampling} The naive iterative decoding results in higher sample diversity but lower fidelity, while the critical resampling results in higher fidelity but lower sample diversity. To maintain the benefits of both approaches while avoiding their drawbacks, the critical reverse sampling bridges the gap between them.
More specifically, the initial token sets are sampled by the naive iterative decoding, however the sets are likely to contain less likely tokens due to the independent sampling, uncorrectable sampling, and the modeling error. 
To tackle this limitation, the critical reverse sampling iteratively removes such less likely tokens by retracting the step from $T$ to $t^*$. 
The resulting masked token sets serve as diverse contexts to produce a wide range of synthetic samples.
We then resample the token sets using the critical resampling, leading to diverse and high-fidelity synthetic samples. 

\section{Experiments}

\subsection{Dataset}
We use all datasets from the UCR Time Series archive \cite{UCRArchive2018} in our experiments. 
The archive contains a diverse set of time series datasets with different characteristics, such as varying lengths, different numbers of classes, and challenging patterns.
The use of this archive for TSG evaluation was first introduced by \cite{lee2023vector}, and was shown effective for a fair evaluation.
In the experiments, each dataset is normalized such that it has zero mean and unit variance. 
The results shown in the following subsections are obtained using datasets containing 100 samples or more, totaling 86 datasets. Full results on all datasets are reported in the appendix. 

\subsection{Evaluation Metrics}
Our experiments primarily use three metrics: Inception Scores (IS), Fréchet Inception Distance (FID) scores, and Classification Accuracy Score (CAS) to evaluate synthetic sample quality. 
IS measures label prediction entropy and evenness across all labels using a pretrained model such as Inception~v3 \cite{szegedy2015going} for IMG and Fully Convolutional Network (FCN) for TSG \cite{wang2017time, lee2023vector}. 
FID compares distributions of generated and real samples with lower scores indicating better quality. 
CAS trains a classifier such as ResNet-50 for IMG and FCN for TSG on synthetic samples and tests it on real samples to measure resulting accuracy.
\cite{lee2023vector} describes the evaluation protocol for TSG in detail.

In the evaluation process, better performance is indicated by lower FID scores and higher IS values. Similarly, a higher value of CAS also suggests improved performance.

\subsection{Experimental Setup}
For our encoder and decoder, we utilize those from VQ-VAE \cite{van2017neural}, and the VQ library from \cite{vq_github} is used for implementation. The bidirectional transformers for the prior models are implemented using code from \cite{x_transformer_github}. 
ESS is applicable to both LF and HF token sampling processes, but we found that it is sufficient to apply it to the LF sampling process only, as the HF sampling is conditional sampling, therefore easier than unconditional sampling like the LF sampling \cite{bao2022conditional}. 
Further information on the implementation and parameter choices can be found in the appendix.


\subsection{Unconditional Time Series Generation}

The experimental results are presented comparatively, considering popular TSG methods such as GMMN \cite{li2015generative}, RCGAN \cite{esteban2017real}, TimeGAN \cite{yoon2019time}, SigCWGAN \cite{ni2020conditional}, and TimeVQVAE. We also include TimeVQVAE with ESS, referred to as TimeVQVAE+ESS, to assess its impact. Additionally, we incorporate TimeVQVAE with Token-Critic, denoted as TimeVQVAE+Token-Critic, to evaluate the effect of Token-Critic compared to ESS within the context of TSG.

\begin{figure}[!ht]
\centering
  \includegraphics[width=0.75\linewidth]{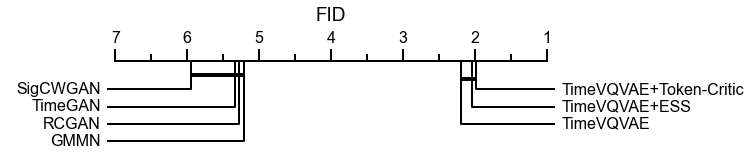}
  \includegraphics[width=0.75\linewidth]{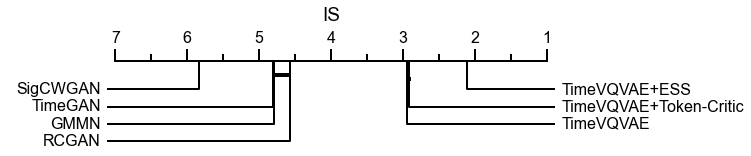}
\caption{CD diagram that compares the different TSG methods in terms of FID score and IS.
The FID scores are multiplied by a factor of -1 to rank them in ascending order, where a higher rank indicates a better overall FID score. 
Conversely, for IS, a higher rank indicates a better overall IS. 
}
\label{fig:CD-unconditional}
\end{figure}

Figure~\ref{fig:CD-unconditional} presents critical diagrams (CDs) that compare the different methods in terms of FID score and IS. It is apparent that there exists a large margin between the GAN-based approaches and the VQ-based approaches. 
It also shows a slightly positive performance gain on TimeVQVAE by Token-Critic in terms of FID. Yet, the performance improvement is the largest on TimeVQVAE with ESS, achieving a significant improvement on IS, demonstrating the effectiveness of its enhanced sampling.

\begin{figure}[!ht]
\centering
\includegraphics[width=0.8\linewidth]{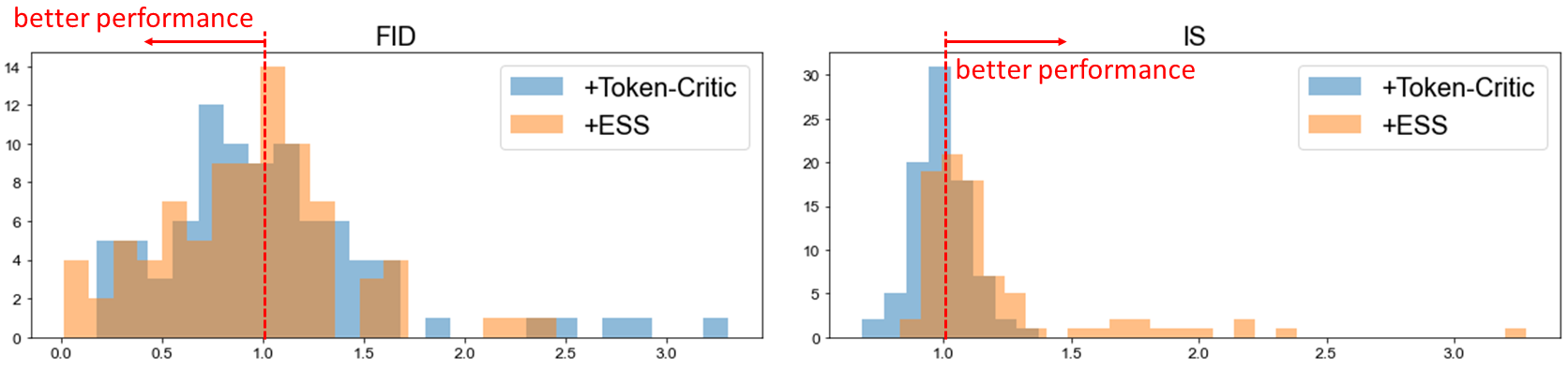}
\caption{Quantitative comparison of performance gain on TimeVQVAE by Token-Critic and ESS. The bar graphs represent histograms where the values are computed as $\frac{\text{score of TimeVQVAE+Token-Critic}}{\text{score of TimeVQVAE}}$ for +Token-Critic, and $\frac{\text{score of TimeVQVAE+ESS}}{\text{score of TimeVQVAE}}$ for +ESS.
The score is either FID score or IS. 
Then, the FID ratio below 1.0 indicates a positive improvement from TimeVQVAE, and the IS ratio above 1.0 indicates a positive improvement.}
\label{fig:quantitative_comp_token-critic_vs_ess}
\end{figure}

\begin{figure}[!ht]
\centering
\includegraphics[width=0.99\linewidth]{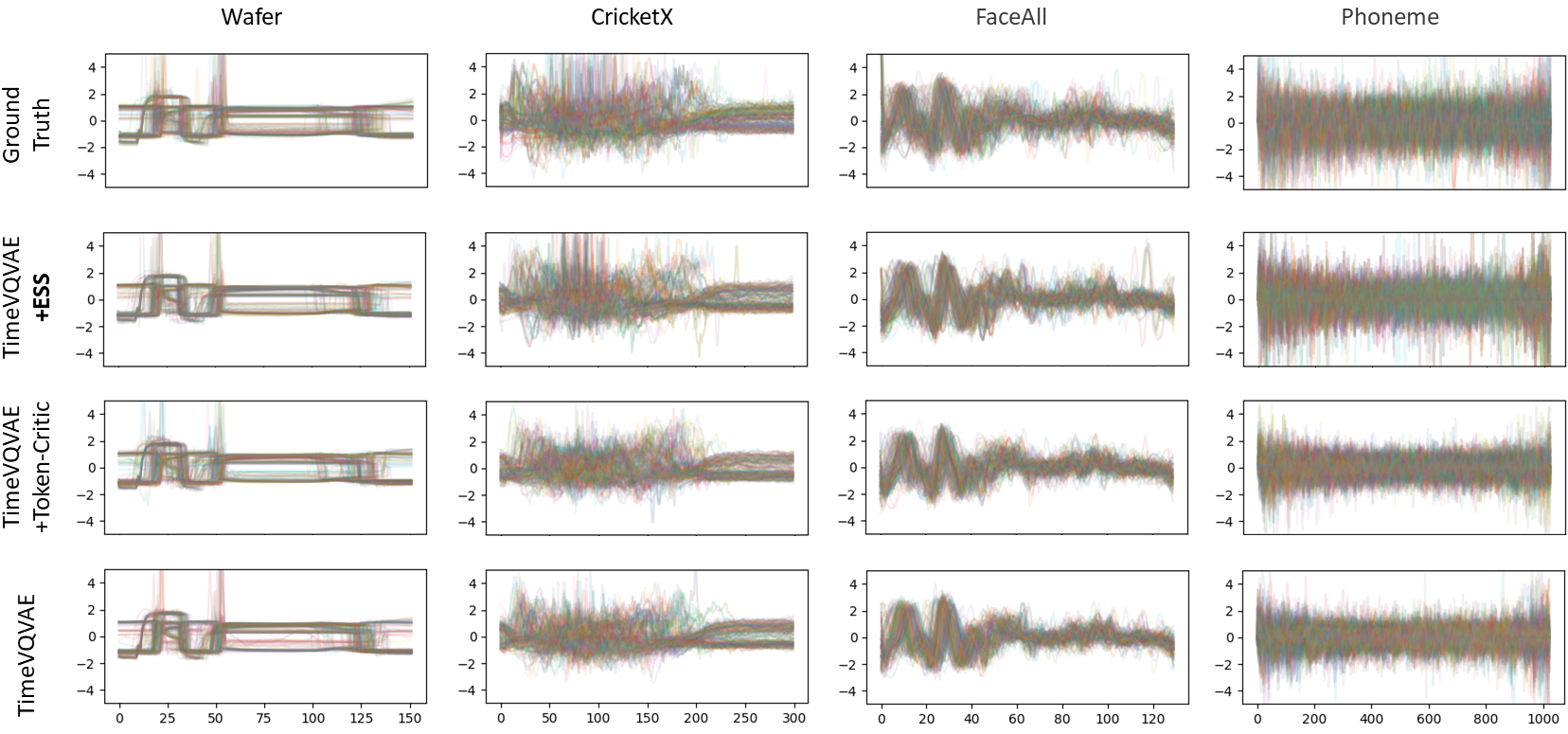}
\caption{Comparative visualization of unconditionally-generated samples by TimeVQVAE and TimeVQVAE with Token-Critic and ESS, respectively. The column names represent dataset names, and the ground truth samples are from the test sets.
A subset of the test set is randomly chosen and visualized for Ground Truth to present the overall distribution.}
\label{fig:unconditional_sampling_visualization}
\end{figure}

Figure~\ref{fig:quantitative_comp_token-critic_vs_ess} depicts a quantitative comparison of performance gain on TimeVQVAE by Token-Critic and ESS. 
Notably, +ESS exhibits a higher number of histogram bars on the right side above 1.0 for IS, suggesting a large improvement from TimeVQVAE. 
The visual comparison between +Token-Critic and +ESS is presented in figure~\ref{fig:unconditional_sampling_visualization}.

\begin{figure}[!ht]
\centering
\includegraphics[width=0.75\linewidth]{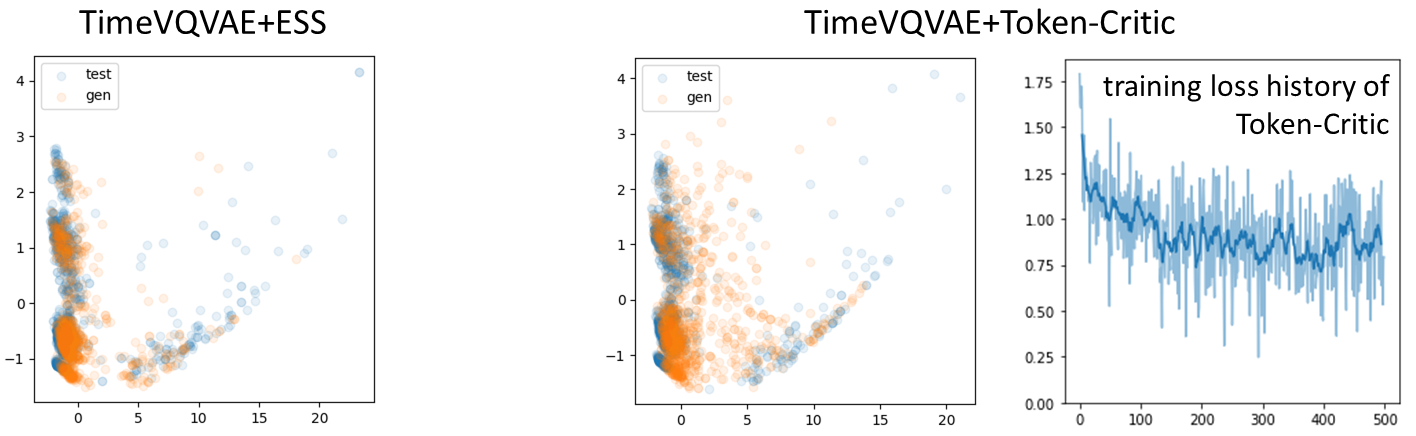}
\caption{Visualization of the representations of two sets of generated samples by TimeVQVAE+ESS and TimeVQVAE+Token-Critic, respectively, along with the training loss history of Token-Critic during its stage~3 training.
It is important to highlight that a binary cross entropy loss above 0.3 is generally indicative of limited learning.
The blue and orange dots denote representations of test and generated samples, respectively. 
The target dataset in this case is Wafer.
}
\label{fig:limitation_token_critic}
\end{figure}

In our experiments, we observed that Token-Critic faces challenges in minimizing the discriminative loss, as shown in figure~\ref{fig:limitation_token_critic}, since tokens sampled with the prior model often resemble tokens from the training set, making them difficult to distinguish. As a result, Token-Critic struggles to accurately capture the realism of the generated tokens. Ironically, this situation leads to higher FID scores due to the insufficient realism measurement, which introduces ambiguity in the sampling process and results in greater sample diversity compared to TimeVQVAE. Consequently, the generated samples cover a larger portion of the distribution. However, this property also limits the precise capture of class-conditional distribution, ultimately leading to a lower value of CAS. 

Figure~\ref{fig:limitation_token_critic} demonstrates Token-Critic's higher sample diversity but imprecise capture of the target distribution. The overlap between the distributions of test and generated samples is evident in TimeVQVAE+ESS, while TimeVQVAE+Token-Critic shows a wider distribution than the test data. This reflects the higher sample diversity but imprecise capture of the target distribution, stemming from the difficulty of minimizing the discriminative loss (binary cross-entropy).


\subsection{Class-conditional Time Series Generation}

\begin{figure}[!ht]
\centering
\includegraphics[width=0.75\linewidth]{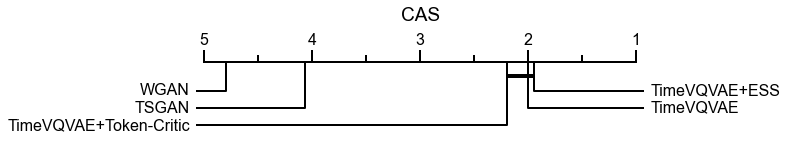}
\includegraphics[width=0.75\linewidth]{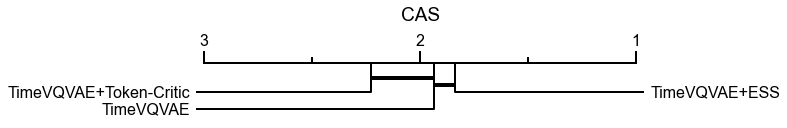}
\caption{CD diagrams that compare the different class-conditional TSG methods in terms of CAS.
The first CD diagram is evaluated on 70 subset datasets from the UCR Time Series archive, for which WGAN and TSGAN are available. 
The second CD diagram is evaluated on all 128 datasets from the archive. 
A higher rank indicates higher CAS overall.
}
\label{fig:CAS}
\end{figure}

TimeVQVAE, TimeVQVAE+Token-Critic, and TimeVQVAE+ESS are evaluated for class-conditional sampling in comparison to the popular class-conditional TSG methods such as TSGAN \cite{smith2020conditional} and WGAN \cite{wgan2017}.
The study conducted by \cite{smith2020conditional} provides the CAS scores for WGAN and TSGAN on 70 subset datasets of the UCR Time Series archive. 
In figure~\ref{fig:CAS}, we compare WGAN, TSGAN, and TimeVQVAE in terms of CAS for class-conditional sampling. We did not utilize the classifier-free guidance proposed in \cite{lee2023vector}, and therefore, the guidance scale was set to 1. 
The results demonstrate that +ESS outperforms the other methods in capturing the class-conditional sample distributions, while +Token-Critic struggles.

\subsection{Ablation Studies}


    



\subsubsection{Performance Gains by Critical Reverse Sampling and Critical Resampling}
\begin{figure}[!ht]
\centering
  \includegraphics[width=0.70\linewidth]{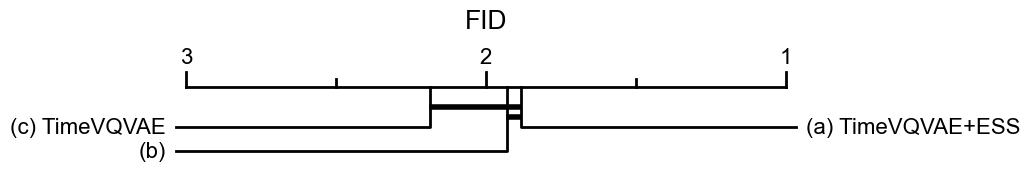}
  \includegraphics[width=0.70\linewidth]{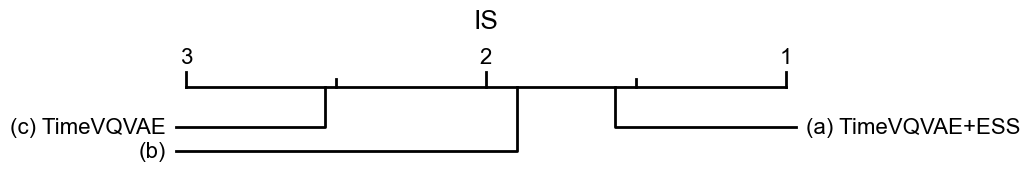}
\caption{CD diagrams that compare performance gains by critical reverse sampling and critical resampling in terms of FID score and IS.
}
\label{fig:CD-ablation-ESS}
\end{figure}

\begin{figure}[!ht]
\centering
\includegraphics[width=0.8\linewidth]{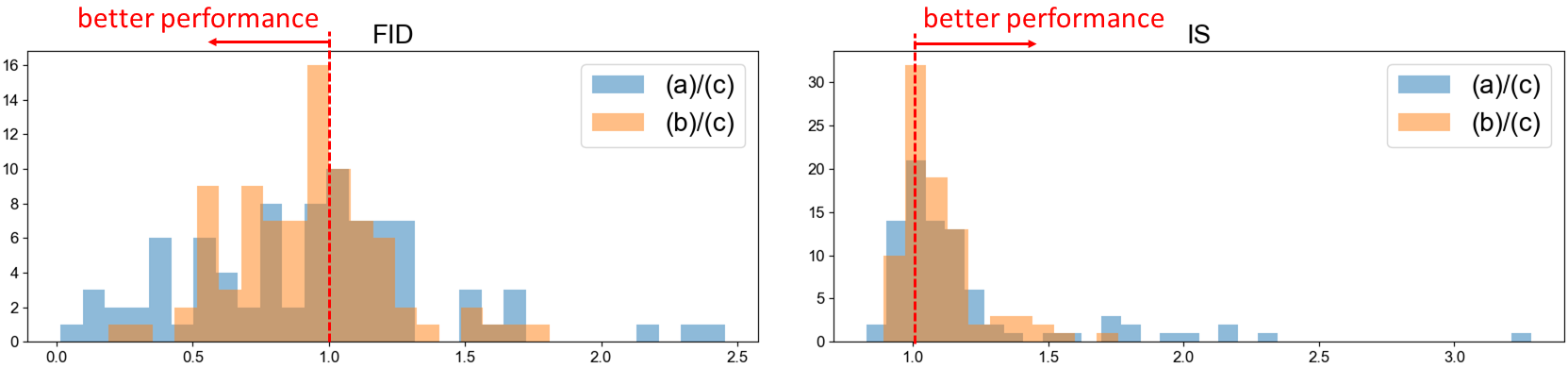}
\caption{Quantitative comparison of performance gains by critical reverse sampling and critical resampling. 
The bar graphs represents histograms where the values are computed as $\frac{\text{score of (a)}}{\text{score of (c)}}$ for (a)/(c), and $\frac{\text{score of (b)}}{\text{score of (c)}}$ for (b)/(c).
The score is either FID score or IS. 
}
\label{fig:ablation-ESS-quantitative}
\end{figure}

We investigate the effect of each component of ESS and measure the performance gains by critical reverse sampling and critical resampling. 
Table~\ref{tab:ESS_respective_performance_gain_ablation_study_table} describes the ablation study cases, figure~\ref{fig:CD-ablation-ESS} presents the CDs that compare the three ablation study cases in terms of FID score and IS, and figure~\ref{fig:ablation-ESS-quantitative} presents a quantitative visualization of the performance gains. 
The results reveal a significant improvement in both IS with the adoption of critical reverse sampling. Moreover, the performance is substantially boosted by incorporating critical resampling, resulting in a substantial leap in overall performance.

\begin{table}[!ht]
    \caption{Ablation study cases with respect to the components of ESS. The signs of \texttt{o} and \texttt{x} indicate the use of the item described in the corresponding column name, where \texttt{o} and \texttt{x} denote using and not using, respectively. 
    The aim of this ablation study is to measure the performance gain by critical reverse sampling and critical resampling, respectively.
    (b) specifically represents iterative decoding $\rightarrow$ critical reverse sampling $\rightarrow$ iterative decoding.}
    \label{tab:ESS_respective_performance_gain_ablation_study_table}
    \centering
    \begin{tabular}{llccc l}
    \toprule
    ~ & Description & \makecell{Naïve\\iterative\\decoding} & \makecell{Critical\\reverse\\sampling} & \makecell{Critical\\resampling} & Remarks \\
    \midrule
    (a) & TimeVQVAE+ESS & \texttt{o} & \texttt{o} & \texttt{o} & ~ \\
    (b) & ~ & \texttt{o} & \texttt{o} & \texttt{x} & \makecell[l]{iterative decoding $\rightarrow$ \\critical reverse sampling $\rightarrow$ \\iterative decoding} \\
    (c) & TimeVQVAE & \texttt{o} & \texttt{x} & \texttt{x} \\ 
    \bottomrule
    \end{tabular}
\end{table}

\subsubsection{Transition of Confidence by self-Token-Critic in the Sampling Process}
In the sampling process, naive iterative decoding produces a sub-optimal token set containing less likely tokens due to its limitations. Critical reverse sampling removes such less likely tokens, and critical resampling allows for resampling more likely tokens.

If we measure the realism of sampled tokens at a range of decoding steps from $0$ to $T$ to $t^*$ to $T^*$ ($0 \rightarrow T$ in naive iterative decoding, $T \rightarrow t^*$ in critical reverse sampling, $t^* \rightarrow T^*$ in critical resampling), we can expect that the realism of sampled token set at $T$ is sub-optimal, while the realism at $t^*$ should be higher than at $t$, and the realism at $T^*$ should be optimal and higher than $T$. 

\begin{figure}[!ht]
\centering
\includegraphics[width=0.999\linewidth]{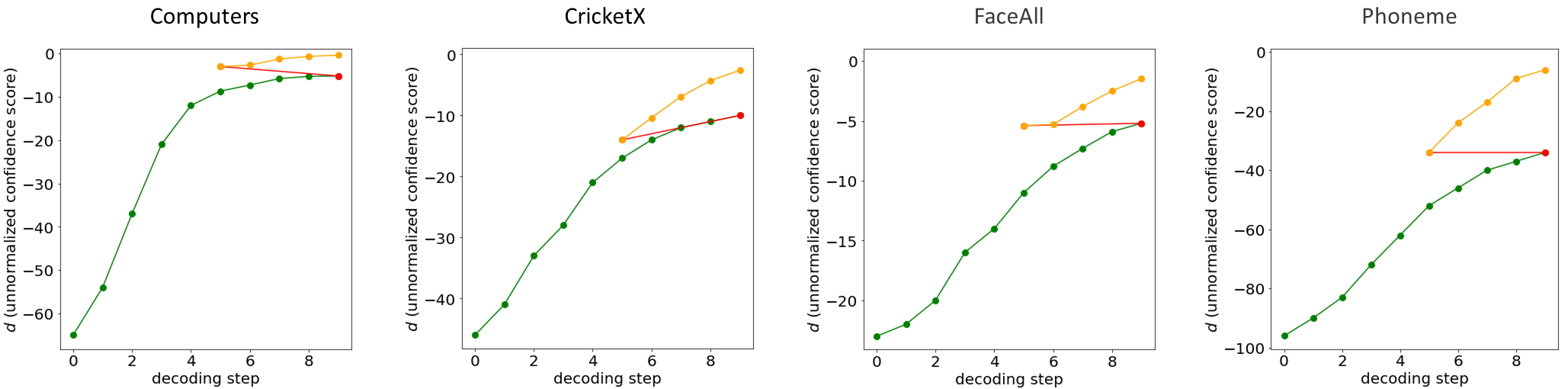}
\caption{Examples of the realism of sampled tokens at a range of decoding steps from $0$ to $T$ (naive iterative decoding, green) to $t^*$ (critical reverse sampling, red) to $T^*$ (critical resampling, yellow). 
The realism is measured using self-Token-Critic.
The column names represent dataset names. 
}
\label{fig:confidence_transition_in_sampling}
\end{figure}

Figure~\ref{fig:confidence_transition_in_sampling} presents examples of the realism of sampled tokens at the range of decoding steps for several datasets, where the realism is measured using the unnormalized confidence score, $d$, from \eqref{eq:confdience_score}. More precisely, $\sum_j d_j$ measures the realism of a sampled token set, representing the y-axis in the figure.


\section{Conclusion}
In this study, we have identified the limitations of the sampling process used in TimeVQVAE, MaskGIT, and Token-Critic, and proposed the ESS to overcome the limitations and achieve effective sampling. 
ESS is a novel sampling scheme that ensures both sample diversity and fidelity, in which naive iterative decoding ensures sample diversity, and critical reverse sampling and critical resampling ensure fidelity.
ESS has demonstrated significant performance gains in both FID scores, IS, and CAS on the UCR Time Series Archive \cite{UCRArchive2018} compared to other TSG methods, indicating its superior generative performance in both unconditional sampling and class-conditional sampling. ESS produces samples with both higher fidelity and greater sample diversity. 
At the same time, we observed that Token-Critic \cite{lezama2022improved} struggles in minimizing the discriminative loss on this collection of datasets. We are unsure why Token-Critic struggles on the datasets we have tested on, but we hypothesize that Token-Critic may be better suited for larger datasets, such as the ones presented in the original paper.
Our ablation studies have additionally demonstrated the effectiveness of critical reverse sampling and critical resampling.
Overall, our research shows that ESS significantly improves sampling performance on a challenging and diverse collection of time series data sets compared to the baseline sampling methods in MaskGIT \cite{chang2022maskgit}, Token-Critic \cite{lezama2022improved}, and TimeVQVAE \cite{lee2023vector}. 

Although ESS has been evaluated and demonstrated on TSG in this work, ESS is not designed to be specific to time series. ESS is applicable to non-autoregressive masked modeling featuring a VQVAE encoder-decoder, such as generative image modeling and audio modeling. It remains an open question if ESS can provide similar performance gains in these domains, and we leave this to a future study.

\subsubsection*{Acknowledgements}
We would like to thank the Norwegian Research Council for funding the Machine Learning for Irregular Time Series (ML4ITS) project (312062). This funding directly supported this research.
We also would like to thank all the people who have contributed to the UCR time series classification archive \cite{UCRArchive2018}.

\subsubsection*{Ethical Statement}
No conflicts of interest were present during the research process.

\clearpage
\bibliographystyle{unsrt}  
\bibliography{references}  

\begin{thebibliography}{10}

\bibitem{kingma2013auto}
Diederik~P Kingma and Max Welling.
\newblock Auto-encoding variational bayes.
\newblock {\em arXiv preprint arXiv:1312.6114}, 2013.

\bibitem{goodfellow2014generative}
Ian Goodfellow, Jean Pouget-Abadie, Mehdi Mirza, Bing Xu, David Warde-Farley,
  Sherjil Ozair, Aaron Courville, and Yoshua Bengio.
\newblock Generative adversarial nets.
\newblock {\em Advances in neural information processing systems}, 27, 2014.

\bibitem{van2017neural}
Aaron Van Den~Oord, Oriol Vinyals, et~al.
\newblock Neural discrete representation learning.
\newblock {\em Advances in neural information processing systems}, 30, 2017.

\bibitem{ho2020denoising}
Jonathan Ho, Ajay Jain, and Pieter Abbeel.
\newblock Denoising diffusion probabilistic models.
\newblock {\em Advances in neural information processing systems},
  33:6840--6851, 2020.

\bibitem{yu2022scaling}
Jiahui Yu, Yuanzhong Xu, Jing~Yu Koh, Thang Luong, Gunjan Baid, Zirui Wang,
  Vijay Vasudevan, Alexander Ku, Yinfei Yang, Burcu~Karagol Ayan, et~al.
\newblock Scaling autoregressive models for content-rich text-to-image
  generation.
\newblock {\em arXiv preprint arXiv:2206.10789}, 2022.

\bibitem{saharia2022photorealistic}
Chitwan Saharia, William Chan, Saurabh Saxena, Lala Li, Jay Whang, Emily~L
  Denton, Kamyar Ghasemipour, Raphael Gontijo~Lopes, Burcu Karagol~Ayan, Tim
  Salimans, et~al.
\newblock Photorealistic text-to-image diffusion models with deep language
  understanding.
\newblock {\em Advances in Neural Information Processing Systems},
  35:36479--36494, 2022.

\bibitem{ramesh2022hierarchical}
Aditya Ramesh, Prafulla Dhariwal, Alex Nichol, Casey Chu, and Mark Chen.
\newblock Hierarchical text-conditional image generation with clip latents.
\newblock {\em arXiv preprint arXiv:2204.06125}, 2022.

\bibitem{borsos2023audiolm}
Zal{\'a}n Borsos, Rapha{\"e}l Marinier, Damien Vincent, Eugene Kharitonov,
  Olivier Pietquin, Matt Sharifi, Dominik Roblek, Olivier Teboul, David
  Grangier, Marco Tagliasacchi, et~al.
\newblock Audiolm: a language modeling approach to audio generation.
\newblock {\em IEEE/ACM Transactions on Audio, Speech, and Language
  Processing}, 2023.

\bibitem{liu2023audioldm}
Haohe Liu, Zehua Chen, Yi~Yuan, Xinhao Mei, Xubo Liu, Danilo Mandic, Wenwu
  Wang, and Mark~D Plumbley.
\newblock Audioldm: Text-to-audio generation with latent diffusion models.
\newblock {\em arXiv preprint arXiv:2301.12503}, 2023.

\bibitem{lee2023vector}
Daesoo Lee, Sara Malacarne, and Erlend Aune.
\newblock Vector quantized time series generation with a bidirectional prior
  model.
\newblock In {\em International Conference on Artificial Intelligence and
  Statistics}, pages 7665--7693. PMLR, 2023.

\bibitem{salimans2022progressive}
Tim Salimans and Jonathan Ho.
\newblock Progressive distillation for fast sampling of diffusion models.
\newblock {\em arXiv preprint arXiv:2202.00512}, 2022.

\bibitem{song2020denoising}
Jiaming Song, Chenlin Meng, and Stefano Ermon.
\newblock Denoising diffusion implicit models.
\newblock {\em arXiv preprint arXiv:2010.02502}, 2020.

\bibitem{chang2022maskgit}
Huiwen Chang, Han Zhang, Lu~Jiang, Ce~Liu, and William~T Freeman.
\newblock Maskgit: Masked generative image transformer.
\newblock In {\em Proceedings of the IEEE/CVF Conference on Computer Vision and
  Pattern Recognition}, pages 11315--11325, 2022.

\bibitem{lezama2022improved}
Jos{\'e} Lezama, Huiwen Chang, Lu~Jiang, and Irfan Essa.
\newblock Improved masked image generation with token-critic.
\newblock In {\em European Conference on Computer Vision}, pages 70--86.
  Springer, 2022.

\bibitem{UCRArchive2018}
Hoang~Anh Dau, Eamonn Keogh, Kaveh Kamgar, Chin-Chia~Michael Yeh, Yan Zhu,
  Shaghayegh Gharghabi, Chotirat~Ann Ratanamahatana, Yanping, Bing Hu, Nurjahan
  Begum, Anthony Bagnall, Abdullah Mueen, Gustavo Batista, and Hexagon-ML.
\newblock The ucr time series classification archive, October 2018.

\bibitem{ang2023tsgbench}
Yihao Ang, Qiang Huang, Yifan Bao, Anthony~KH Tung, and Zhiyong Huang.
\newblock Tsgbench: Time series generation benchmark.
\newblock {\em arXiv preprint arXiv:2309.03755}, 2023.

\bibitem{esteban2017real}
Crist{\'o}bal Esteban, Stephanie~L Hyland, and Gunnar R{\"a}tsch.
\newblock Real-valued (medical) time series generation with recurrent
  conditional gans.
\newblock {\em arXiv preprint arXiv:1706.02633}, 2017.

\bibitem{yoon2019time}
Jinsung Yoon, Daniel Jarrett, and Mihaela Van~der Schaar.
\newblock Time-series generative adversarial networks.
\newblock {\em Advances in neural information processing systems}, 32, 2019.

\bibitem{ni2020conditional}
Hao Ni, Lukasz Szpruch, Magnus Wiese, Shujian Liao, and Baoren Xiao.
\newblock Conditional sig-wasserstein gans for time series generation.
\newblock {\em arXiv preprint arXiv:2006.05421}, 2020.

\bibitem{smith2020conditional}
Kaleb~E Smith and Anthony~O Smith.
\newblock Conditional gan for timeseries generation.
\newblock {\em arXiv preprint arXiv:2006.16477}, 2020.

\bibitem{li2022tts}
Xiaomin Li, Anne Hee~Hiong Ngu, and Vangelis Metsis.
\newblock Tts-cgan: A transformer time-series conditional gan for biosignal
  data augmentation.
\newblock {\em arXiv preprint arXiv:2206.13676}, 2022.

\bibitem{szegedy2015going}
Christian Szegedy, Wei Liu, Yangqing Jia, Pierre Sermanet, Scott Reed, Dragomir
  Anguelov, Dumitru Erhan, Vincent Vanhoucke, and Andrew Rabinovich.
\newblock Going deeper with convolutions.
\newblock In {\em Proceedings of the IEEE conference on computer vision and
  pattern recognition}, pages 1--9, 2015.

\bibitem{wang2017time}
Zhiguang Wang, Weizhong Yan, and Tim Oates.
\newblock Time series classification from scratch with deep neural networks: A
  strong baseline.
\newblock In {\em 2017 International joint conference on neural networks
  (IJCNN)}, pages 1578--1585. IEEE, 2017.

\bibitem{vq_github}
Phil Wang, Kenny Olsen, and Wes Bouaziz.
\newblock {lucidrains/vector-quantize-pytorch}.
\newblock {https://github.com/lucidrains/vector-quantize-pytorch}, 2022.

\bibitem{x_transformer_github}
Phil Wang.
\newblock {lucidrains/x-transformers}.
\newblock {https://github.com/lucidrains/x-transformers}, 2022.

\bibitem{bao2022conditional}
Fan Bao, Chongxuan Li, Jiacheng Sun, and Jun Zhu.
\newblock Why are conditional generative models better than unconditional ones?
\newblock {\em arXiv preprint arXiv:2212.00362}, 2022.

\bibitem{li2015generative}
Yujia Li, Kevin Swersky, and Rich Zemel.
\newblock Generative moment matching networks.
\newblock In {\em International conference on machine learning}, pages
  1718--1727. PMLR, 2015.

\bibitem{wgan2017}
Martin Arjovsky, Soumith Chintala, and L{\'e}on Bottou.
\newblock {W}asserstein generative adversarial networks.
\newblock In Doina Precup and Yee~Whye Teh, editors, {\em Proceedings of the
  34th International Conference on Machine Learning}, volume~70 of {\em
  Proceedings of Machine Learning Research}, pages 214--223. PMLR, 06--11 Aug
  2017.

\bibitem{ml4its_timevqvae}
{danelee2601}.
\newblock {ML4ITS/TimeVQVAE}.
\newblock {GitHub}, 2023.
\newblock {https://github.com/ml4its/timevqvae}.

\bibitem{deng2009imagenet}
Jia Deng, Wei Dong, Richard Socher, Li-Jia Li, Kai Li, and Li~Fei-Fei.
\newblock Imagenet: A large-scale hierarchical image database.
\newblock In {\em 2009 IEEE conference on computer vision and pattern
  recognition}, pages 248--255. Ieee, 2009.

\bibitem{maskgit_github}
dome272.
\newblock {dome272/MaskGIT-pytorch}.
\newblock {https://github.com/dome272/MaskGIT-pytorch}, 2022.

\bibitem{smith2020one}
Kaleb~Earl Smith.
\newblock {\em One Dimensional Neural Time Series Generation}.
\newblock PhD thesis, Florida Institute of Technology, 2020.

\end{thebibliography}
\vfill

\pagebreak
\appendix

\section{Implementation Details}

\subsection{Evaluation Metrics: IS, FID, and CAS}
To compute IS and FID, we generate the same number of synthetic samples as that of the test set, except for cases where the test set has less than 1024 samples. In these cases, we generate 1024 synthetic samples to ensure a more representative distribution and improve the consistency of the IS and FID scores. To compute CAS, we generate the same number of synthetic samples per class as that of the training set. However, if the training set has less than 1,000 samples, we generate 1,000 samples according to the class distribution of the training set. This approach ensures the distribution of class-conditionally generated samples and mitigates the overfitting issue in the FCN model's training. For example, if a training set has 50 and 150 samples for class~1 and class~2, respectively, we generate 250 samples for class 1 and 750 samples for class~2, resulting in a total of 1,000 generated samples.
Additionally, the accuracy reported in this paper is the highest test accuracy achieved during the training process. This approach serves to alleviate challenges associated with model regularization and to yield more consistent CAS results. While the final test accuracy may fluctuate based on various regularization parameters, the maximum test accuracy tends to be more stable. This is because it usually aligns with the point of minimum loss, effectively capturing the CAS performance at this optimal stage of training.

\subsection{STFT, ISTFT, Encoder, Decoder} The STFT and ISTFT configurations used in this paper are the same as those employed in TimeVQVAE \cite{lee2023vector}. Additionally, the encoder and decoder configurations are identical to the \textit{Base} models described in the TimeVQVAE paper \cite{lee2023vector} to ensure the comparability of our results with TimeVQVAE. Our implementations of encoder and decoder are directly from \cite{ml4its_timevqvae}.

\subsection{Vector Quantizer}
Our implementation for VQ-VAE is from \cite{vq_github,ml4its_timevqvae}. 
\cite{lee2023vector} set the codebook size $K$ to 32 with the code dimension size being the same as the hidden dimension size of the encoder and decoder for both LF and HF components. We found that the codebook size can be further decreased to 16 since the datasets from the UCR archive are less complex and smaller in scale compared to image benchmark datasets such as ImageNet \cite{deng2009imagenet}. In fact, the downsized codebook eases the prior learning because a prior model is trained with a cross entropy loss, solving a classification task in stage~2 -- \textit{i.e.,} prediction of a likely token among the tokens in the codebook given a mask token. The decrease of the codebook size is equivalent to the decrease of tokens to choose from. 
Hence, we suggest reducing the codebook size to ease the prior learning process, provided that it does not adversely affect the reconstruction quality in stage~1.

\subsection{Prior Learning}
The number of iterations, $T$, is set to 10, and the cosine masking scheduling function is used, following \cite{chang2022maskgit,lee2023vector}.
Our implementation for MaskGIT is adopted from \cite{maskgit_github,lee2023vector} and implementation for the prior models, \textit{i.e.,} bidirectional transformers, is from \cite{x_transformer_github}. \cite{lee2023vector} introduces the different sizes for the prior model, and we use the \textit{Base} model to let our results comparable to TimeVQVAE.

\subsection{Optimizer}
The AdamW optimizer is used with \{batch size for stage~1: 128, batch size for stage~2: 256, initial learning rate: 1e-3, learning rate scheduler: cosine scheduler, weight decay: 1e-5\}. 
The maximum epochs are \{stage~1: 2,000, stage~2: 10,000\} in the unconditional and class-conditional sampling experiments.

\section{Full Results}
\label{appendix:results}

The full result tables for the evaluation metrics such as FID, IS, and CAS are reported below.

\begingroup
\footnotesize
\begin{longtable}{lrrrrrrr} 
\caption{Full FID score results, where a lower score indicates better performance.
The FID score larger than 1,000 is marked as nan because such a large FID score already indicates that the synthetic sample quality is significantly poor.}
\label{tab:full_restuls_fid}
\\
\toprule
Dataset names & \makecell[c]{GMMN \\ \cite{lee2023vector}} & \makecell[c]{RCGAN \\ \cite{lee2023vector}} & \makecell[c]{TimeGAN \\ \cite{lee2023vector}} & \makecell[c]{SigCWGAN \\ \cite{lee2023vector}} & TimeVQVAE & \makecell[c]{TimeVQVAE\\+Token-Critic} & \makecell[c]{TimeVQVAE\\+ESS} \\ 
\midrule
ACSF1 & 74.3 & 98.3 & 81.3 & nan & 28.0 & 18.9 & 24.8  \\ 
Adiac & 98.2 & 54.3 & 45.1 & nan & 9.5 & 7.5 & 9.1  \\ 
AllGestureWiimoteX & 180.7 & nan & nan & nan & 1.6 & 0.8 & 1.9  \\ 
AllGestureWiimoteY & 23.5 & 0.0 & nan & 660.4 & 1.6 & 1.0 & 2.6  \\ 
AllGestureWiimoteZ & nan & 2.8 & 7.9 & 278.9 & 2.4 & 0.9 & 2.4  \\ 
ArrowHead & 8.1 & 15.5 & 22.2 & nan & 2.5 & 1.8 & 2.7  \\ 
BME & 77.6 & 179.6 & 140.1 & nan & 50.7 & 46.3 & 38.7  \\ 
Beef & 215.2 & 36.0 & 18.3 & nan & 2.6 & 1.9 & 2.6  \\ 
BeetleFly & 246.4 & 348.5 & nan & nan & 4.9 & 4.8 & 3.4  \\ 
BirdChicken & 9.9 & 8.5 & 7.6 & 735.5 & 0.2 & 0.2 & 0.4  \\ 
CBF & 147.1 & 483.9 & 5.5 & 24.4 & 2.5 & 2.0 & 4.4  \\ 
Car & 101.5 & 579.5 & 186.1 & nan & 3.9 & 3.7 & 4.1  \\ 
Chinatown & 27.3 & 70.4 & 59.9 & 25.8 & 2.4 & 2.0 & 2.4  \\ 
ChlorineConcentration & 41.6 & 7.4 & 7.1 & 132.4 & 0.7 & 0.6 & 0.7  \\ 
CinCECGTorso & nan & 172.3 & 57.0 & nan & 13.2 & 13.1 & 11.5  \\ 
Coffee & 16.2 & 18.1 & nan & nan & 0.1 & 0.1 & 0.3  \\ 
Computers & 29.1 & nan & nan & 19.8 & 3.8 & 4.7 & 2.2  \\ 
CricketX & 115.8 & 404.8 & 23.0 & 26.9 & 3.6 & 3.9 & 1.4  \\ 
CricketY & 33.0 & 0.0 & nan & 17.6 & 3.1 & 3.0 & 1.7  \\ 
CricketZ & 60.1 & 105.3 & 14.4 & 19.8 & 3.6 & 4.1 & 2.1  \\ 
Crop & 80.7 & 14.9 & 18.7 & 35.3 & 3.0 & 2.4 & 5.1  \\ 
DiatomSizeReduction & 138.6 & 232.9 & 88.7 & nan & 11.9 & 9.9 & 15.3  \\ 
DistalPhalanxOutlineAgeGroup & 13.9 & 9.0 & 110.6 & nan & 7.7 & 1.4 & 9.7  \\ 
DistalPhalanxOutlineCorrect & 9.2 & 12.0 & 16.0 & 50.4 & 0.8 & 0.3 & 1.9  \\ 
DistalPhalanxTW & 16.8 & 19.2 & 21.3 & nan & 14.6 & 3.8 & 15.9  \\ 
DodgerLoopDay & 56.8 & 429.9 & 30.1 & 14.4 & 2.2 & 1.8 & 6.0  \\ 
DodgerLoopGame & 44.8 & nan & 21.6 & 63.4 & 5.3 & 4.4 & 7.5  \\ 
DodgerLoopWeekend & 9.6 & 273.2 & 16.1 & 14.7 & 5.4 & 4.9 & 6.0  \\ 
ECG200 & 3.0 & 2.8 & 2.7 & 20.6 & 1.2 & 0.9 & 1.3  \\ 
ECG5000 & 26.6 & 4.5 & 35.2 & 55.9 & 0.9 & 0.8 & 0.6  \\ 
ECGFiveDays & 15.0 & 22.3 & 7.1 & 523.2 & 4.2 & 3.3 & 4.0  \\ 
EOGHorizontalSignal & 287.8 & 47.3 & 95.5 & nan & 3.1 & 3.1 & 3.9  \\ 
EOGVerticalSignal & 238.1 & 457.7 & 100.6 & nan & 6.9 & 5.6 & 7.2  \\ 
Earthquakes & nan & nan & nan & 5.4 & 1.8 & 1.9 & 0.6  \\ 
ElectricDevices & 37.1 & 151.9 & 79.9 & 105.5 & 8.7 & 13.6 & 9.4  \\ 
EthanolLevel & 19.4 & 15.7 & 18.0 & nan & 0.3 & 0.3 & 0.4  \\ 
FaceAll & 42.4 & 10.3 & nan & 33.0 & 4.6 & 6.4 & 3.1  \\ 
FaceFour & 18.6 & 65.0 & 56.5 & 26.5 & 4.0 & 2.7 & 6.7  \\ 
FacesUCR & 39.7 & 7.3 & 20.6 & 39.7 & 2.4 & 3.7 & 1.1  \\ 
FiftyWords & nan & 27.7 & 81.7 & 821.9 & 12.7 & 20.5 & 3.7  \\ 
Fish & nan & 36.3 & 47.4 & nan & 12.0 & 11.1 & 11.3  \\ 
FordA & 3.6 & 178.0 & nan & nan & 3.0 & 3.0 & 4.5  \\ 
FordB & nan & 45.6 & nan & nan & 1.2 & 2.8 & 1.1  \\ 
FreezerRegularTrain & 56.0 & 41.6 & 27.6 & 176.9 & 8.2 & 7.8 & 8.8  \\ 
FreezerSmallTrain & 49.3 & 32.0 & 42.7 & 109.6 & 8.9 & 8.8 & 8.5  \\ 
Fungi & 48.4 & 86.0 & 82.9 & nan & 4.9 & 5.0 & 4.5  \\ 
GestureMidAirD1 & nan & nan & nan & nan & 3.7 & 4.5 & 3.0  \\ 
GestureMidAirD2 & nan & nan & 306.8 & nan & 89.0 & 72.3 & 95.6  \\ 
GestureMidAirD3 & 214.7 & nan & nan & nan & 5.5 & 6.5 & 5.3  \\ 
GesturePebbleZ1 & 163.4 & 29.5 & 10.6 & 26.2 & 1.0 & 3.3 & 1.0  \\ 
GesturePebbleZ2 & 16.7 & 350.8 & 22.3 & 49.9 & 1.7 & 1.1 & 1.0  \\ 
GunPoint & 16.0 & 8.8 & 3.5 & nan & 0.6 & 1.0 & 0.8  \\ 
GunPointAgeSpan & 314.7 & 19.7 & 267.3 & nan & 0.6 & 0.7 & 0.5  \\ 
GunPointMaleVersusFemale & nan & 44.6 & 237.5 & 215.4 & 0.4 & 1.1 & 0.3  \\ 
GunPointOldVersusYoung & 14.1 & nan & 13.2 & nan & 0.6 & 0.3 & 0.4  \\ 
Ham & nan & 43.5 & 27.4 & 427.0 & 0.5 & 0.4 & 0.5  \\ 
HandOutlines & 9.4 & 3.7 & 1.3 & nan & 0.2 & 0.6 & 0.3  \\ 
Haptics & nan & nan & nan & nan & 2.8 & 2.3 & 2.8  \\ 
Herring & 35.0 & 3.9 & 75.8 & nan & 0.6 & 0.6 & 0.6  \\ 
HouseTwenty & 26.6 & nan & 38.0 & 100.3 & 2.8 & 1.7 & 2.9  \\ 
InlineSkate & 96.0 & 143.3 & 119.5 & nan & 18.5 & 18.0 & 24.2  \\ 
InsectEPGRegularTrain & 7.3 & 28.5 & 0.9 & 0.3 & 3.5 & 2.9 & 4.8  \\ 
InsectEPGSmallTrain & 121.2 & 113.5 & 125.9 & 80.8 & 0.7 & 0.1 & 1.1  \\ 
InsectWingbeatSound & 14.5 & 6.2 & 132.3 & nan & 12.5 & 17.6 & 1.7  \\ 
ItalyPowerDemand & 5.7 & 57.5 & 5.8 & 4.6 & 2.6 & 1.7 & 3.3  \\ 
LargeKitchenAppliances & 8.3 & 47.3 & 31.8 & 281.0 & 0.8 & 1.1 & 0.5  \\ 
Lightning2 & 19.5 & 8.6 & 71.7 & 162.3 & 2.1 & 0.9 & 1.4  \\ 
Lightning7 & 79.2 & 78.6 & 27.0 & 92.1 & 1.2 & 1.1 & 2.3  \\ 
Mallat & nan & 11.4 & nan & nan & 1.7 & 1.4 & 1.7  \\ 
Meat & 134.7 & 18.2 & nan & nan & 5.5 & 4.8 & 5.2  \\ 
MedicalImages & 20.9 & 24.0 & nan & 55.3 & 3.2 & 2.4 & 2.4  \\ 
MelbournePedestrian & 66.6 & 35.5 & 62.5 & 90.1 & 2.9 & 1.2 & 6.3  \\ 
MiddlePhalanxOutlineAgeGroup & 18.7 & 12.6 & 13.8 & 590.2 & 18.3 & 10.5 & 22.4  \\ 
MiddlePhalanxOutlineCorrect & 7.2 & 21.2 & 162.3 & 535.4 & 1.4 & 0.4 & 1.6  \\ 
MiddlePhalanxTW & 62.2 & 27.4 & 19.8 & nan & 5.1 & 1.8 & 6.3  \\ 
MixedShapesRegularTrain & 379.0 & 412.2 & nan & nan & 43.9 & 58.0 & 15.8  \\ 
MixedShapesSmallTrain & 20.6 & 13.3 & 262.1 & nan & 10.3 & 12.8 & 4.1  \\ 
MoteStrain & 5.6 & 5.9 & 4.0 & 13.9 & 2.1 & 2.7 & 3.1  \\ 
NonInvasiveFetalECGThorax1 & 438.5 & nan & nan & nan & 21.4 & 16.1 & 23.0  \\ 
NonInvasiveFetalECGThorax2 & 126.8 & 117.9 & 150.0 & nan & 32.5 & 25.7 & 36.3  \\ 
OSULeaf & nan & nan & nan & nan & 19.9 & 22.3 & 7.2  \\ 
OliveOil & 9.8 & 9.5 & 9.9 & nan & 1.8 & 1.7 & 1.8  \\ 
PLAID & 797.4 & nan & 335.9 & nan & 370.4 & 107.5 & 281.7  \\ 
PhalangesOutlinesCorrect & 1.9 & 3.0 & 2.3 & 73.4 & 0.5 & 0.3 & 0.8  \\ 
Phoneme & nan & nan & nan & nan & 12.4 & 14.5 & 4.3  \\ 
PickupGestureWiimoteZ & 275.6 & 380.9 & nan & 116.0 & 6.6 & 2.5 & 4.3  \\ 
PigAirwayPressure & 111.2 & 188.5 & 642.6 & 274.4 & 50.7 & 55.1 & 59.0  \\ 
PigArtPressure & 291.4 & nan & nan & 80.3 & 138.2 & 144.6 & 27.3  \\ 
PigCVP & 107.6 & nan & nan & 343.3 & 58.2 & 50.2 & 62.1  \\ 
Plane & 39.2 & 38.9 & 36.2 & nan & 5.2 & 3.7 & 4.8  \\ 
PowerCons & 18.3 & 15.0 & 16.5 & 30.0 & 0.9 & 0.8 & 0.9  \\ 
ProximalPhalanxOutlineAgeGroup & 73.2 & 93.9 & 22.1 & nan & 0.7 & 1.9 & 0.8  \\ 
ProximalPhalanxOutlineCorrect & 3.7 & 1.7 & 13.8 & nan & 0.5 & 0.1 & 0.8  \\ 
ProximalPhalanxTW & 60.8 & 24.2 & 16.2 & nan & 3.3 & 1.2 & 8.1  \\ 
RefrigerationDevices & nan & nan & nan & 9.2 & 22.0 & 23.5 & 2.7  \\ 
Rock & 0.0 & 0.0 & 0.0 & 23.5 & 2.6 & 4.3 & 2.6  \\ 
ScreenType & 32.7 & nan & 30.9 & 28.4 & 6.7 & 6.9 & 5.3  \\ 
SemgHandGenderCh2 & 14.4 & nan & nan & 69.9 & 4.5 & 4.9 & 3.5  \\ 
SemgHandMovementCh2 & 164.6 & 71.8 & 199.1 & 59.4 & 12.7 & 8.8 & 16.2  \\ 
SemgHandSubjectCh2 & 37.0 & nan & nan & 108.1 & 32.5 & 26.2 & 31.4  \\ 
ShakeGestureWiimoteZ & nan & 42.5 & 22.4 & 116.7 & 1.0 & 1.2 & 0.8  \\ 
ShapeletSim & 8.4 & 2.6 & nan & 0.8 & 6.7 & 6.1 & 11.2  \\ 
ShapesAll & nan & nan & nan & nan & 14.9 & 12.4 & 19.3  \\ 
SmallKitchenAppliances & 19.0 & nan & 20.8 & 233.1 & 5.5 & 8.4 & 6.2  \\ 
SmoothSubspace & nan & 7.7 & 14.1 & 8.2 & 3.4 & 3.4 & 2.6  \\ 
SonyAIBORobotSurface1 & 31.7 & nan & 14.3 & 81.7 & 13.8 & 12.0 & 12.3  \\ 
SonyAIBORobotSurface2 & 22.6 & 21.2 & 14.5 & 10.7 & 1.9 & 0.9 & 2.0  \\ 
StarLightCurves & 27.6 & 42.9 & 6.7 & nan & 1.1 & 0.6 & 1.9  \\ 
Strawberry & 70.6 & 20.8 & 333.9 & nan & 0.2 & 0.3 & 0.4  \\ 
SwedishLeaf & 46.3 & 16.5 & 24.7 & 151.5 & 8.4 & 6.4 & 8.4  \\ 
Symbols & 33.8 & 29.9 & 57.2 & 407.6 & 6.2 & 1.4 & 4.9  \\ 
SyntheticControl & 14.7 & 12.6 & 14.0 & 17.4 & 2.7 & 4.8 & 1.5  \\ 
ToeSegmentation1 & 19.2 & 502.3 & nan & 7.0 & 2.1 & 2.6 & 0.8  \\ 
ToeSegmentation2 & 3.5 & 2.8 & 154.7 & 89.0 & 5.8 & 3.9 & 4.4  \\ 
Trace & 93.0 & 90.8 & 21.7 & 187.4 & 5.3 & 3.1 & 6.3  \\ 
TwoLeadECG & 10.4 & 12.2 & 8.7 & 49.5 & 0.4 & 0.5 & 0.2  \\ 
TwoPatterns & 15.4 & 31.8 & 29.6 & 51.3 & 3.7 & 5.5 & 1.4  \\ 
UMD & 454.1 & 11.4 & 15.5 & 501.3 & 1.3 & 0.8 & 0.9  \\ 
UWaveGestureLibraryAll & 737.2 & nan & nan & 586.1 & 4.4 & 5.3 & 3.2  \\ 
UWaveGestureLibraryX & 31.5 & nan & nan & nan & 7.5 & 11.7 & 7.0  \\ 
UWaveGestureLibraryY & nan & 22.0 & nan & nan & 7.7 & 11.2 & 4.9  \\ 
UWaveGestureLibraryZ & 9.6 & 24.3 & nan & nan & 41.7 & 59.4 & 23.9  \\ 
Wafer & 26.8 & 0.0 & 23.6 & 164.0 & 1.8 & 1.5 & 1.5  \\ 
Wine & nan & nan & nan & nan & 0.5 & 0.9 & 0.4  \\ 
WordSynonyms & 34.4 & 11.3 & 42.3 & nan & 4.0 & 3.8 & 4.6  \\ 
Worms & nan & nan & nan & nan & 8.3 & 10.4 & 1.9  \\ 
WormsTwoClass & nan & nan & nan & nan & 7.1 & 7.5 & 0.8  \\ 
Yoga & nan & nan & nan & nan & 6.4 & 10.8 & 0.1  \\ 
\bottomrule
\end{longtable}
\endgroup

\begingroup
\footnotesize
\begin{longtable}{lrrrrrrr} 
\caption{Full IS results, where a higher score indicates better performance.
}
\label{tab:full_restuls_is}
\\
\toprule
Dataset names & \makecell[c]{GMMN \\ \cite{lee2023vector}} & \makecell[c]{RCGAN \\ \cite{lee2023vector}} & \makecell[c]{TimeGAN \\ \cite{lee2023vector}} & \makecell[c]{SigCWGAN \\ \cite{lee2023vector}} & TimeVQVAE & \makecell[c]{TimeVQVAE\\+Token-Critic} & \makecell[c]{TimeVQVAE\\+ESS} \\ 
\midrule
ACSF1 & 1.4 & 1.9 & 1.5 & 2.0 & 3.4 & 4.6 & 3.9  \\ 
Adiac & 1.1 & 1.4 & 1.6 & 1.6 & 5.7 & 7.2 & 6.0  \\ 
AllGestureWiimoteX & 3.6 & 3.2 & 4.3 & 1.5 & 2.1 & 2.2 & 2.2  \\ 
AllGestureWiimoteY & 3.4 & 0.0 & 3.1 & 1.7 & 1.9 & 2.0 & 1.8  \\ 
AllGestureWiimoteZ & 2.6 & 2.3 & 2.3 & 1.0 & 1.0 & 1.1 & 1.0  \\ 
ArrowHead & 1.9 & 1.1 & 1.1 & 1.0 & 2.6 & 2.5 & 2.7  \\ 
BME & 1.0 & 1.2 & 1.2 & 1.1 & 1.8 & 1.9 & 1.7  \\ 
Beef & 1.2 & 1.4 & 1.5 & 1.0 & 2.6 & 2.5 & 2.9  \\ 
BeetleFly & 1.0 & 1.1 & 1.0 & 1.0 & 1.8 & 1.9 & 2.0  \\ 
BirdChicken & 1.0 & 1.1 & 1.1 & 1.0 & 1.9 & 1.9 & 1.9  \\ 
CBF & 1.0 & 1.1 & 1.5 & 1.1 & 2.7 & 2.7 & 2.9  \\ 
Car & 1.7 & 1.0 & 1.0 & 1.0 & 2.2 & 2.2 & 2.3  \\ 
Chinatown & 1.8 & 1.0 & 1.3 & 1.6 & 1.9 & 1.9 & 1.9  \\ 
ChlorineConcentration & 1.2 & 2.0 & 1.9 & 1.0 & 1.9 & 1.9 & 1.9  \\ 
CinCECGTorso & 1.5 & 1.3 & 1.6 & 1.0 & 1.4 & 1.4 & 1.7  \\ 
Coffee & 1.1 & 1.0 & 1.0 & 1.0 & 1.9 & 1.8 & 1.9  \\ 
Computers & 1.2 & 1.8 & 1.4 & 1.2 & 1.5 & 1.5 & 1.7  \\ 
CricketX & 1.8 & 2.0 & 2.3 & 1.8 & 2.9 & 2.7 & 6.3  \\ 
CricketY & 2.3 & 0.0 & 3.6 & 2.0 & 3.6 & 3.3 & 6.9  \\ 
CricketZ & 3.0 & 1.9 & 2.5 & 1.9 & 3.1 & 2.8 & 6.8  \\ 
Crop & 5.7 & 7.7 & 7.8 & 7.5 & 17.6 & 16.9 & 18.2  \\ 
DiatomSizeReduction & 1.0 & 1.2 & 1.3 & 1.0 & 3.1 & 3.0 & 2.8  \\ 
DistalPhalanxOutlineAgeGroup & 1.5 & 1.6 & 1.3 & 1.0 & 1.9 & 2.3 & 1.7  \\ 
DistalPhalanxOutlineCorrect & 1.2 & 1.4 & 1.7 & 1.8 & 1.6 & 1.8 & 1.4  \\ 
DistalPhalanxTW & 1.9 & 1.9 & 1.7 & 2.1 & 2.6 & 3.1 & 2.4  \\ 
DodgerLoopDay & 1.0 & 1.2 & 1.2 & 1.3 & 3.2 & 2.7 & 5.6  \\ 
DodgerLoopGame & 1.0 & 1.2 & 1.0 & 1.1 & 1.9 & 1.7 & 1.9  \\ 
DodgerLoopWeekend & 1.2 & 1.0 & 1.6 & 1.1 & 1.9 & 1.9 & 2.0  \\ 
ECG200 & 1.4 & 1.5 & 1.4 & 1.5 & 1.6 & 1.7 & 1.6  \\ 
ECG5000 & 1.7 & 1.7 & 1.6 & 1.6 & 2.0 & 2.0 & 2.0  \\ 
ECGFiveDays & 1.1 & 1.0 & 1.2 & 1.7 & 1.7 & 1.6 & 1.7  \\ 
EOGHorizontalSignal & 1.3 & 2.5 & 1.3 & 2.0 & 4.7 & 4.6 & 5.3  \\ 
EOGVerticalSignal & 1.5 & 1.7 & 1.7 & 1.2 & 3.3 & 3.4 & 3.7  \\ 
Earthquakes & 1.2 & 1.0 & 1.0 & 1.1 & 1.1 & 1.1 & 1.2  \\ 
ElectricDevices & 2.9 & 3.8 & 4.4 & 2.8 & 4.5 & 4.3 & 4.5  \\ 
EthanolLevel & 1.0 & 1.1 & 1.0 & 2.0 & 1.2 & 1.3 & 1.2  \\ 
FaceAll & 2.0 & 2.2 & 3.0 & 1.8 & 5.3 & 3.8 & 9.6  \\ 
FaceFour & 1.2 & 1.3 & 1.1 & 1.2 & 2.6 & 3.0 & 3.3  \\ 
FacesUCR & 1.0 & 1.9 & 2.1 & 1.5 & 4.2 & 2.9 & 8.6  \\ 
FiftyWords & 2.7 & 1.9 & 1.3 & 1.8 & 4.7 & 3.8 & 8.4  \\ 
Fish & 1.0 & 1.2 & 1.2 & 1.0 & 3.1 & 2.8 & 3.8  \\ 
FordA & 1.5 & 1.3 & 1.0 & 1.0 & 1.5 & 1.5 & 1.4  \\ 
FordB & 1.0 & 1.7 & 1.0 & 1.0 & 1.5 & 1.5 & 1.5  \\ 
FreezerRegularTrain & 1.0 & 1.0 & 1.0 & 1.6 & 1.4 & 1.5 & 1.4  \\ 
FreezerSmallTrain & 1.8 & 2.0 & 1.9 & 1.2 & 1.9 & 1.9 & 1.9  \\ 
Fungi & 1.0 & 1.1 & 1.3 & 1.0 & 5.8 & 5.4 & 8.0  \\ 
GestureMidAirD1 & 2.1 & 2.5 & 2.6 & 1.4 & 2.5 & 2.9 & 2.4  \\ 
GestureMidAirD2 & 1.1 & 1.1 & 1.5 & 1.0 & 2.1 & 2.4 & 2.0  \\ 
GestureMidAirD3 & 3.3 & 2.9 & 1.7 & 2.6 & 1.9 & 1.8 & 1.9  \\ 
GesturePebbleZ1 & 1.2 & 1.1 & 2.0 & 1.4 & 1.2 & 1.2 & 1.3  \\ 
GesturePebbleZ2 & 1.3 & 1.7 & 1.9 & 1.0 & 1.4 & 1.6 & 1.5  \\ 
GunPoint & 1.4 & 1.3 & 1.6 & 1.0 & 1.8 & 1.8 & 1.9  \\ 
GunPointAgeSpan & 1.5 & 1.1 & 1.0 & 1.0 & 1.7 & 1.7 & 1.8  \\ 
GunPointMaleVersusFemale & 1.0 & 1.1 & 1.0 & 1.0 & 1.9 & 1.9 & 2.0  \\ 
GunPointOldVersusYoung & 1.7 & 1.3 & 1.9 & 1.3 & 1.9 & 2.0 & 2.0  \\ 
Ham & 1.1 & 1.5 & 1.4 & 1.3 & 1.7 & 1.6 & 1.8  \\ 
HandOutlines & 1.2 & 1.0 & 1.4 & 1.0 & 1.2 & 1.4 & 1.2  \\ 
Haptics & 1.9 & 1.0 & 1.0 & 1.0 & 1.9 & 1.9 & 2.0  \\ 
Herring & 1.0 & 1.0 & 1.1 & 1.0 & 1.5 & 1.3 & 1.5  \\ 
HouseTwenty & 1.0 & 1.2 & 1.0 & 1.0 & 1.6 & 1.7 & 1.7  \\ 
InlineSkate & 1.7 & 1.2 & 1.7 & 1.0 & 1.6 & 1.5 & 1.8  \\ 
InsectEPGRegularTrain & 2.5 & 2.2 & 2.8 & 2.8 & 2.3 & 2.3 & 2.3  \\ 
InsectEPGSmallTrain & 1.0 & 1.0 & 1.0 & 1.0 & 2.6 & 2.6 & 2.5  \\ 
InsectWingbeatSound & 2.3 & 1.8 & 1.1 & 1.5 & 3.0 & 2.6 & 5.4  \\ 
ItalyPowerDemand & 1.6 & 1.2 & 1.6 & 1.6 & 2.0 & 2.0 & 2.0  \\ 
LargeKitchenAppliances & 1.4 & 1.5 & 1.4 & 1.8 & 2.3 & 2.2 & 2.7  \\ 
Lightning2 & 1.2 & 1.2 & 1.5 & 1.0 & 1.5 & 1.7 & 1.6  \\ 
Lightning7 & 1.0 & 1.1 & 1.6 & 1.4 & 3.3 & 3.6 & 5.1  \\ 
Mallat & 1.0 & 1.1 & 1.0 & 1.0 & 4.1 & 4.0 & 4.4  \\ 
Meat & 1.1 & 1.2 & 1.4 & 1.0 & 1.4 & 1.4 & 1.4  \\ 
MedicalImages & 2.4 & 1.2 & 1.4 & 2.4 & 2.6 & 2.9 & 3.3  \\ 
MelbournePedestrian & 4.2 & 4.8 & 3.8 & 4.7 & 8.8 & 8.7 & 8.3  \\ 
MiddlePhalanxOutlineAgeGroup & 1.6 & 1.6 & 1.7 & 1.1 & 2.0 & 2.2 & 1.9  \\ 
MiddlePhalanxOutlineCorrect & 1.4 & 1.7 & 1.1 & 1.1 & 1.6 & 1.7 & 1.5  \\ 
MiddlePhalanxTW & 2.2 & 2.0 & 2.3 & 2.4 & 2.7 & 2.8 & 2.6  \\ 
MixedShapesRegularTrain & 2.2 & 1.9 & 2.3 & 1.2 & 1.4 & 1.2 & 2.3  \\ 
MixedShapesSmallTrain & 1.3 & 2.2 & 1.5 & 1.0 & 2.2 & 2.1 & 2.7  \\ 
MoteStrain & 1.8 & 1.5 & 1.4 & 1.2 & 1.9 & 1.9 & 2.0  \\ 
NonInvasiveFetalECGThorax1 & 2.3 & 1.4 & 2.9 & 1.0 & 7.3 & 7.6 & 7.9  \\ 
NonInvasiveFetalECGThorax2 & 2.0 & 1.2 & 2.6 & 1.6 & 7.2 & 6.7 & 7.7  \\ 
OSULeaf & 2.1 & 1.6 & 1.0 & 1.1 & 1.4 & 1.2 & 2.3  \\ 
OliveOil & 1.2 & 1.0 & 1.0 & 1.0 & 1.0 & 1.0 & 1.0  \\ 
PLAID & 1.9 & 2.8 & 2.2 & 1.3 & 1.7 & 1.9 & 1.7  \\ 
PhalangesOutlinesCorrect & 1.3 & 1.5 & 1.5 & 1.0 & 1.4 & 1.5 & 1.4  \\ 
Phoneme & 3.7 & 3.7 & 3.0 & 2.0 & 2.7 & 2.2 & 8.9  \\ 
PickupGestureWiimoteZ & 1.3 & 1.5 & 1.2 & 1.0 & 1.2 & 1.3 & 1.1  \\ 
PigAirwayPressure & 2.5 & 1.2 & 1.0 & 1.2 & 3.4 & 3.1 & 3.8  \\ 
PigArtPressure & 3.6 & 2.9 & 2.1 & 1.2 & 2.0 & 1.8 & 4.6  \\ 
PigCVP & 3.0 & 1.6 & 1.6 & 1.1 & 2.8 & 3.2 & 3.3  \\ 
Plane & 1.2 & 1.6 & 1.3 & 1.0 & 5.3 & 5.3 & 5.9  \\ 
PowerCons & 1.2 & 1.1 & 1.1 & 1.5 & 1.8 & 1.7 & 1.9  \\ 
ProximalPhalanxOutlineAgeGroup & 1.0 & 1.7 & 1.9 & 1.9 & 2.5 & 2.5 & 2.4  \\ 
ProximalPhalanxOutlineCorrect & 1.4 & 1.6 & 1.4 & 1.2 & 1.5 & 1.7 & 1.5  \\ 
ProximalPhalanxTW & 1.5 & 2.2 & 1.8 & 1.1 & 2.6 & 2.9 & 2.1  \\ 
RefrigerationDevices & 1.3 & 1.2 & 2.0 & 1.3 & 1.4 & 1.3 & 2.4  \\ 
Rock & 0.0 & 0.0 & 0.0 & 1.0 & 2.5 & 2.5 & 2.6  \\ 
ScreenType & 1.4 & 1.2 & 1.2 & 1.3 & 2.1 & 2.0 & 2.3  \\ 
SemgHandGenderCh2 & 1.0 & 1.3 & 1.0 & 1.0 & 1.5 & 1.4 & 1.4  \\ 
SemgHandMovementCh2 & 2.2 & 1.8 & 1.4 & 1.1 & 1.9 & 2.0 & 1.9  \\ 
SemgHandSubjectCh2 & 2.0 & 3.6 & 3.1 & 1.1 & 1.9 & 1.9 & 2.0  \\ 
ShakeGestureWiimoteZ & 1.9 & 2.5 & 2.4 & 1.3 & 1.3 & 1.4 & 1.4  \\ 
ShapeletSim & 1.1 & 1.1 & 1.2 & 1.1 & 1.7 & 1.6 & 1.9  \\ 
ShapesAll & 1.4 & 3.2 & 2.7 & 1.3 & 6.2 & 5.5 & 8.5  \\ 
SmallKitchenAppliances & 1.3 & 1.4 & 1.2 & 1.6 & 1.6 & 1.4 & 1.8  \\ 
SmoothSubspace & 1.0 & 1.9 & 1.9 & 1.8 & 2.5 & 2.4 & 2.5  \\ 
SonyAIBORobotSurface1 & 1.1 & 1.1 & 1.3 & 1.0 & 1.5 & 1.6 & 1.6  \\ 
SonyAIBORobotSurface2 & 1.3 & 1.2 & 1.5 & 1.5 & 1.8 & 1.9 & 1.9  \\ 
StarLightCurves & 1.8 & 1.8 & 1.6 & 1.1 & 2.3 & 2.2 & 2.5  \\ 
Strawberry & 1.3 & 1.0 & 1.0 & 1.0 & 1.8 & 1.8 & 1.8  \\ 
SwedishLeaf & 1.6 & 2.5 & 2.0 & 1.6 & 7.3 & 6.8 & 9.2  \\ 
Symbols & 1.3 & 1.1 & 2.2 & 1.3 & 4.2 & 5.0 & 4.8  \\ 
SyntheticControl & 2.4 & 2.3 & 2.4 & 2.1 & 4.6 & 4.2 & 5.4  \\ 
ToeSegmentation1 & 1.0 & 1.4 & 1.7 & 1.3 & 1.8 & 1.9 & 1.9  \\ 
ToeSegmentation2 & 1.2 & 1.2 & 1.6 & 1.0 & 1.7 & 1.8 & 1.9  \\ 
Trace & 1.1 & 1.0 & 1.5 & 1.0 & 3.3 & 3.4 & 3.3  \\ 
TwoLeadECG & 1.3 & 1.3 & 1.4 & 1.2 & 1.9 & 1.9 & 2.0  \\ 
TwoPatterns & 1.6 & 1.7 & 1.6 & 1.6 & 2.5 & 2.2 & 3.1  \\ 
UMD & 1.0 & 1.8 & 1.0 & 1.3 & 2.6 & 2.5 & 2.7  \\ 
UWaveGestureLibraryAll & 2.4 & 2.1 & 3.2 & 1.9 & 3.2 & 2.9 & 4.0  \\ 
UWaveGestureLibraryX & 2.3 & 2.2 & 1.0 & 2.2 & 3.9 & 3.1 & 4.4  \\ 
UWaveGestureLibraryY & 3.2 & 1.8 & 1.0 & 1.8 & 3.3 & 2.8 & 3.9  \\ 
UWaveGestureLibraryZ & 2.6 & 2.9 & 2.2 & 1.1 & 2.8 & 2.2 & 3.4  \\ 
Wafer & 1.2 & 0.0 & 1.1 & 1.6 & 1.3 & 1.4 & 1.3  \\ 
Wine & 1.1 & 1.8 & 1.1 & 1.0 & 1.6 & 1.5 & 1.5  \\ 
WordSynonyms & 1.3 & 1.6 & 1.4 & 1.4 & 1.8 & 1.7 & 2.0  \\ 
Worms & 1.2 & 1.5 & 1.1 & 1.1 & 2.4 & 2.1 & 2.7  \\ 
WormsTwoClass & 1.2 & 1.8 & 1.5 & 1.5 & 1.1 & 1.0 & 1.6  \\ 
Yoga & 1.5 & 1.1 & 1.0 & 1.0 & 1.3 & 1.2 & 1.6  \\ 
\bottomrule
\end{longtable}
\endgroup

\begingroup
\footnotesize
\centering
\begin{longtable}{lrrrrrrr} 
\caption{Full CAS results, where a higher score indicates better performance.
}
\label{tab:full_restuls_cas}
\\
\toprule
Dataset names & WGAN \cite{smith2020one} & TSGAN \cite{smith2020one} & TimeVQVAE & \makecell{TimeVQVAE\\+Token-Critic} & \makecell{TimeVQVAE\\+ESS} \\ 
\midrule
ACSF1 & ~ & ~ & 88.0 & 82.0 & 85.0  \\ 
        Adiac & ~ & ~ & 77.0 & 77.8 & 77.7  \\ 
        AllGestureWiimoteX & ~ & ~ & 87.9 & 82.4 & 85.2  \\ 
        AllGestureWiimoteY & ~ & ~ & 85.1 & 84.0 & 84.0  \\ 
        AllGestureWiimoteZ & ~ & ~ & 78.1 & 79.7 & 81.3  \\ 
        ArrowHead & 61.7 & 85.7 & 85.7 & 86.3 & 86.3  \\ 
        BME & 80.0 & 82.7 & 72.0 & 74.7 & 74.0  \\ 
        Beef & 20.0 & 60.0 & 70.0 & 70.0 & 66.7  \\ 
        BeetleFly & 55.0 & 90.0 & 95.0 & 95.0 & 95.0  \\ 
        BirdChicken & 90.0 & 75.0 & 100.0 & 100.0 & 100.0  \\ 
        CBF & 70.9 & 87.3 & 100.0 & 100.0 & 100.0  \\ 
        Car & 65.0 & 70.0 & 91.7 & 86.7 & 90.0  \\ 
        Chinatown & ~ & ~ & 100.0 & 100.0 & 100.0  \\ 
        ChlorineConcentration & 56.5 & 54.5 & 70.3 & 67.6 & 70.3  \\ 
        CinCECGTorso & 40.6 & 49.1 & 73.0 & 78.1 & 69.9  \\ 
        Coffee & 100.0 & 100.0 & 100.0 & 100.0 & 100.0  \\ 
        Computers & 50.8 & 65.2 & 82.0 & 84.4 & 83.2  \\ 
        CricketX & ~ & ~ & 75.0 & 77.3 & 79.9  \\ 
        CricketY & ~ & ~ & 71.9 & 75.8 & 79.3  \\ 
        CricketZ & ~ & ~ & 82.8 & 76.9 & 82.8  \\ 
        Crop & ~ & ~ & 100.0 & 100.0 & 100.0  \\ 
        DiatomSizeReduction & 73.5 & 79.7 & 98.0 & 98.0 & 98.0  \\ 
        DistalPhalanxOutlineAgeGroup & 73.4 & 70.5 & 75.5 & 77.7 & 75.5  \\ 
        DistalPhalanxOutlineCorrect & 68.8 & 64.9 & 85.0 & 80.0 & 85.0  \\ 
        DistalPhalanxTW & ~ & ~ & 75.5 & 74.8 & 74.8  \\ 
        DodgerLoopDay & ~ & ~ & 56.3 & 57.5 & 55.0  \\ 
        DodgerLoopGame & ~ & ~ & 76.8 & 76.8 & 76.8  \\ 
        DodgerLoopWeekend & ~ & ~ & 97.1 & 98.6 & 96.4  \\ 
        ECG200 & 82.0 & 82.0 & 91.0 & 91.0 & 90.0  \\ 
        ECG5000 & 78.2 & 86.0 & 100.0 & 100.0 & 100.0  \\ 
        ECGFiveDays & 92.6 & 98.8 & 98.8 & 97.7 & 94.6  \\ 
        EOGHorizontalSignal & ~ & ~ & 58.5 & 60.4 & 63.2  \\ 
        EOGVerticalSignal & ~ & ~ & 69.8 & 65.1 & 67.0  \\ 
        Earthquakes & 69.1 & 74.8 & 74.8 & 74.8 & 77.0  \\ 
        ElectricDevices & ~ & ~ & 99.6 & 98.8 & 99.2  \\ 
        EthanolLevel & 24.8 & 30.2 & 34.0 & 30.9 & 36.1  \\ 
        FaceAll & ~ & ~ & 100.0 & 100.0 & 100.0  \\ 
        FaceFour & 65.2 & 84.2 & 94.3 & 94.3 & 94.3  \\ 
        FacesUCR & ~ & ~ & 100.0 & 100.0 & 100.0  \\ 
        FiftyWords & ~ & ~ & 67.8 & 68.3 & 66.8  \\ 
        Fish & ~ & ~ & 95.4 & 93.1 & 96.0  \\ 
        FordA & 80.0 & 89.2 & 95.0 & 93.0 & 95.0  \\ 
        FordB & 62.2 & 61.4 & 88.1 & 69.1 & 83.3  \\ 
        FreezerRegularTrain & 50.0 & 50.1 & 100.0 & 100.0 & 100.0  \\ 
        FreezerSmallTrain & 50.7 & 76.0 & 100.0 & 99.2 & 100.0  \\ 
        Fungi & ~ & ~ & 98.9 & 98.9 & 98.9  \\ 
        GestureMidAirD1 & ~ & ~ & 73.8 & 73.1 & 70.8  \\ 
        GestureMidAirD2 & ~ & ~ & 67.7 & 69.2 & 66.2  \\ 
        GestureMidAirD3 & ~ & ~ & 39.2 & 34.6 & 37.7  \\ 
        GesturePebbleZ1 & ~ & ~ & 90.1 & 89.0 & 89.0  \\ 
        GesturePebbleZ2 & ~ & ~ & 90.5 & 91.8 & 91.1  \\ 
        GunPoint & 100.0 & 100.0 & 100.0 & 100.0 & 100.0  \\ 
        GunPointAgeSpan & 41.3 & 44.1 & 100.0 & 100.0 & 100.0  \\ 
        GunPointMaleVersusFemale & 52.5 & 52.5 & 100.0 & 100.0 & 100.0  \\ 
        GunPointOldVersusYoung & 11.4 & 52.4 & 100.0 & 100.0 & 100.0  \\ 
        Ham & 68.6 & 68.6 & 81.9 & 78.1 & 78.1  \\ 
        HandOutlines & 64.1 & 65.7 & 82.4 & 77.3 & 82.0  \\ 
        Haptics & 20.8 & 31.5 & 53.1 & 54.7 & 53.5  \\ 
        Herring & 46.9 & 65.6 & 71.9 & 70.3 & 71.9  \\ 
        HouseTwenty & 58.0 & 42.1 & 96.6 & 95.8 & 94.1  \\ 
        InlineSkate & ~ & ~ & 47.4 & 39.5 & 44.7  \\ 
        InsectEPGRegularTrain & 64.3 & 64.3 & 100.0 & 100.0 & 100.0  \\ 
        InsectEPGSmallTrain & 16.9 & 35.7 & 100.0 & 100.0 & 100.0  \\ 
        InsectWingbeatSound & ~ & ~ & 49.6 & 52.0 & 48.8  \\ 
        ItalyPowerDemand & ~ & ~ & 100.0 & 100.0 & 100.0  \\ 
        LargeKitchenAppliances & 59.5 & 74.9 & 96.6 & 100.0 & 100.0  \\ 
        Lightning2 & 70.5 & 72.2 & 78.7 & 78.7 & 80.3  \\ 
        Lightning7 & ~ & ~ & 83.6 & 82.2 & 84.9  \\ 
        Mallat & ~ & ~ & 100.0 & 100.0 & 97.6  \\ 
        Meat & 88.3 & 46.7 & 71.7 & 71.7 & 76.7  \\ 
        MedicalImages & ~ & ~ & 75.8 & 77.0 & 80.1  \\ 
        MelbournePedestrian & ~ & ~ & 100.0 & 99.6 & 100.0  \\ 
        MiddlePhalanxOutlineAgeGroup & 40.9 & 50.0 & 66.2 & 61.7 & 65.6  \\ 
        MiddlePhalanxOutlineCorrect & 73.5 & 75.3 & 91.4 & 91.4 & 97.1  \\ 
        MiddlePhalanxTW & ~ & ~ & 61.0 & 59.7 & 61.7  \\ 
        MixedShapesRegularTrain & ~ & ~ & 100.0 & 100.0 & 100.0  \\ 
        MixedShapesSmallTrain & 55.5 & 82.6 & 100.0 & 100.0 & 100.0  \\ 
        MoteStrain & 87.9 & 87.3 & 97.3 & 94.5 & 96.5  \\ 
        NonInvasiveFetalECGThorax1 & ~ & ~ & 63.7 & 68.8 & 66.5  \\ 
        NonInvasiveFetalECGThorax2 & ~ & ~ & 85.5 & 84.8 & 85.2  \\ 
        OSULeaf & ~ & ~ & 88.0 & 87.6 & 96.7  \\ 
        OliveOil & 40.0 & 40.0 & 40.0 & 40.0 & 40.0  \\ 
        PLAID & ~ & ~ & 56.0 & 52.0 & 48.4  \\ 
        PhalangesOutlinesCorrect & 73.9 & 76.9 & 91.0 & 86.7 & 88.7  \\ 
        Phoneme & ~ & ~ & 61.7 & 56.6 & 58.6  \\ 
        PickupGestureWiimoteZ & ~ & ~ & 80.0 & 76.0 & 76.0  \\ 
        PigAirwayPressure & ~ & ~ & 36.1 & 37.5 & 37.5  \\ 
        PigArtPressure & ~ & ~ & 95.2 & 93.8 & 96.6  \\ 
        PigCVP & ~ & ~ & 41.8 & 39.4 & 38.9  \\ 
        Plane & ~ & ~ & 100.0 & 100.0 & 100.0  \\ 
        PowerCons & 51.1 & 52.2 & 93.3 & 92.8 & 95.6  \\ 
        ProximalPhalanxOutlineAgeGroup & 87.3 & 85.9 & 88.3 & 87.8 & 87.3  \\ 
        ProximalPhalanxOutlineCorrect & 84.5 & 88.3 & 97.1 & 94.3 & 94.3  \\ 
        ProximalPhalanxTW & ~ & ~ & 82.0 & 82.0 & 82.4  \\ 
        RefrigerationDevices & 33.1 & 42.4 & 78.2 & 89.9 & 71.1  \\ 
        Rock & 20.0 & 36.0 & 56.0 & 54.0 & 58.0  \\ 
        ScreenType & 52.0 & 56.8 & 58.0 & 100.0 & 66.0  \\ 
        SemgHandGenderCh2 & 65.2 & 65.2 & 100.0 & 100.0 & 100.0  \\ 
        SemgHandMovementCh2 & ~ & ~ & 56.7 & 70.6 & 70.1  \\ 
        SemgHandSubjectCh2 & 26.0 & 29.3 & 54.6 & 56.2 & 53.6  \\ 
        ShakeGestureWiimoteZ & ~ & ~ & 94.0 & 94.0 & 96.0  \\ 
        ShapeletSim & 50.0 & 50.0 & 100.0 & 100.0 & 100.0  \\ 
        ShapesAll & ~ & ~ & 87.5 & 80.9 & 92.0  \\ 
        SmallKitchenAppliances & 56.5 & 64.0 & 95.0 & 95.8 & 100.0  \\ 
        SmoothSubspace & 66.7 & 68.0 & 99.3 & 99.3 & 98.7  \\ 
        SonyAIBORobotSurface1 & 89.5 & 92.8 & 100.0 & 100.0 & 100.0  \\ 
        SonyAIBORobotSurface2 & 92.2 & 84.7 & 98.4 & 98.8 & 99.2  \\ 
        StarLightCurves & 48.7 & 81.4 & 99.6 & 99.2 & 99.6  \\ 
        Strawberry & 83.2 & 93.2 & 97.4 & 98.2 & 98.2  \\ 
        SwedishLeaf & ~ & ~ & 98.8 & 98.4 & 98.8  \\ 
        Symbols & ~ & ~ & 98.8 & 98.4 & 99.6  \\ 
        SyntheticControl & ~ & ~ & 100.0 & 100.0 & 100.0  \\ 
        ToeSegmentation1 & 88.6 & 93.4 & 95.6 & 96.1 & 97.4  \\ 
        ToeSegmentation2 & 80.8 & 91.5 & 90.8 & 90.8 & 90.8  \\ 
        Trace & 97.0 & 100.0 & 100.0 & 100.0 & 100.0  \\ 
        TwoLeadECG & 99.5 & 99.8 & 100.0 & 100.0 & 100.0  \\ 
        TwoPatterns & 76.7 & 86.8 & 96.1 & 95.7 & 94.9  \\ 
        UMD & 97.2 & 97.9 & 99.3 & 99.3 & 99.3  \\ 
        UWaveGestureLibraryAll & ~ & ~ & 78.1 & 76.2 & 82.4  \\ 
        UWaveGestureLibraryX & ~ & ~ & 75.0 & 76.2 & 77.0  \\ 
        UWaveGestureLibraryY & ~ & ~ & 67.2 & 64.1 & 71.9  \\ 
        UWaveGestureLibraryZ & ~ & ~ & 74.2 & 74.4 & 81.3  \\ 
        Wafer & 91.5 & 79.2 & 100.0 & 100.0 & 100.0  \\ 
        Wine & 55.6 & 61.1 & 50.0 & 72.2 & 68.5  \\ 
        WordSynonyms & ~ & ~ & 57.9 & 55.9 & 59.0  \\ 
        Worms & 44.2 & 59.7 & 63.6 & 59.7 & 74.0  \\ 
        WormsTwoClass & 71.4 & 76.6 & 74.0 & 76.6 & 80.5  \\ 
        Yoga & 53.6 & 84.5 & 78.5 & 72.3 & 79.7 \\
\bottomrule
\end{longtable}
\endgroup

\end{document}